\documentclass[10pt,twocolumn,letterpaper]{article}

\usepackage[pagenumbers]{iccv} %
\usepackage{times}
\usepackage{epsfig}
\usepackage{graphicx}
\usepackage{amsmath}
\usepackage{amssymb}
\usepackage{bbm}
\usepackage{tikzducks}
\usepackage{xfp,graphicx}

\usepackage[dvipsnames]{xcolor}

\newcommand{\bA}{\mathbf{A}}

\newcommand{\bF}{\mathbf{F}} %

\newcommand{\bK}{\mathbf{K}}

\newcommand{\bM}{\mathbf{M}}

\newcommand{\bP}{\mathbf{P}}
\newcommand{\bQ}{\mathbf{Q}}

\newcommand{\bs}{\mathbf{s}}

\newcommand{\bV}{\mathbf{V}}

\newcommand{\cX}{\mathcal{X}}

\newcommand{\figref}[1]{Figure~\ref{#1}}
\newcommand{\secref}[1]{Section~\ref{#1}}

\newcommand{\eqnref}[1]{Eq.~\ref{#1}}
\newcommand{\tabref}[1]{Table~\ref{#1}}

\DeclareMathOperator*{\argmin}{arg\,min}

\makeatletter
\DeclareRobustCommand\onedot{\futurelet\@let@token\@onedot}
\def\@onedot{\ifx\@let@token.\else.\null\fi\xspace}
\def\eg{e.g\onedot} 
\def\ie{i.e\onedot}

\def\vs{vs\onedot}

\makeatother

\definecolor{yellow}{rgb}{1,1, 0.6}
\definecolor{lightyellow}{rgb}{1,1, 0.8}
\definecolor{orange}{rgb}{1, 0.8, 0.6}
\definecolor{red}{rgb}{1, 0.6, 0.6}

\definecolor{wincolor}{rgb}{0.85, 0.0, 0.0}

\definecolor{darkyellow}{rgb}{0.8, 0.8, 0.5}
\definecolor{darkred}{rgb}{0.7, 0.3, 0.3}
\definecolor{darkgreen}{rgb}{0.3, 0.7, 0.3}
\definecolor{blue}{rgb}{0, 0, 1.0}
\definecolor{green}{rgb}{0, 1.0, 0}
\definecolor{pink}{rgb}{1, 0.4, 0.7}

\newcommand{\easier}{\mbox{{Easi3R}}\xspace}

\newcommand{\duster}{\mbox{DUSt3R}\xspace}
\newcommand{\master}{\mbox{MASt3R}\xspace}
\newcommand{\monster}{\mbox{MonST3R}\xspace}
\newcommand{\cuter}{\mbox{CUT3R}\xspace}
\newcommand{\daser}{\mbox{DAS3R}\xspace}

\newcommand{\boldparagraph}[1]{\vspace{0.1cm}\noindent{\bf #1}}

\newcommand{\rmnum}[1]{\romannumeral #1}

\definecolor{customlightgray}{rgb}{0.95, 0.95, 0.95} %
\definecolor{darkgreen}{rgb}{0.0, 0.65, 0.0}
\definecolor{darkred}{rgb}{0.75, 0.0, 0.0} %
\definecolor{darkyellow}{rgb}{0.9, 0.72, 0} %
\definecolor{lightyellow}{rgb}{1, 1, 0.8}

\definecolor{DeltaColor}{rgb}{0.039,0.73,0.71}
\definecolor{SigmaColor}{rgb}{0.98,0.45,0.0}
\definecolor{AlphaColor}{rgb}{0,0,0.8}
\definecolor{BetaColor}{rgb}{0.8,0,0.8}
\definecolor{GammaColor}{rgb}{0.514,0.34,0.224}
\definecolor{EpsilonColor}{rgb}{0.353,0.725,0.906}
\definecolor{PurpleColor}{HTML}{9839ff}
\definecolor{RedColor}{rgb}{0.949,0.275, 0.224}
\definecolor{citecolor}{HTML}{0071bc}

\newcommand{\firstcol}{``\raisebox{-0.2em}{\includegraphics[width=1em]{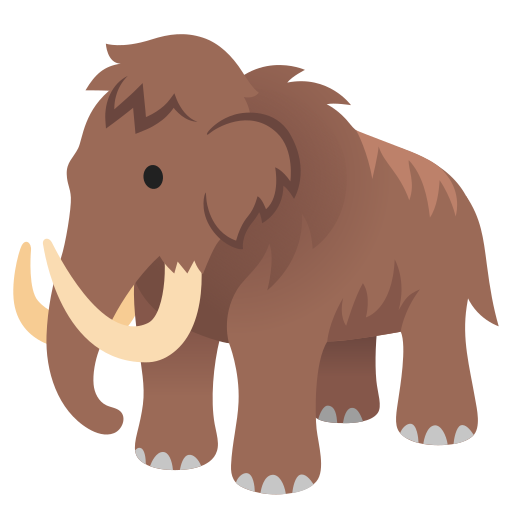} \includegraphics[width=1em]{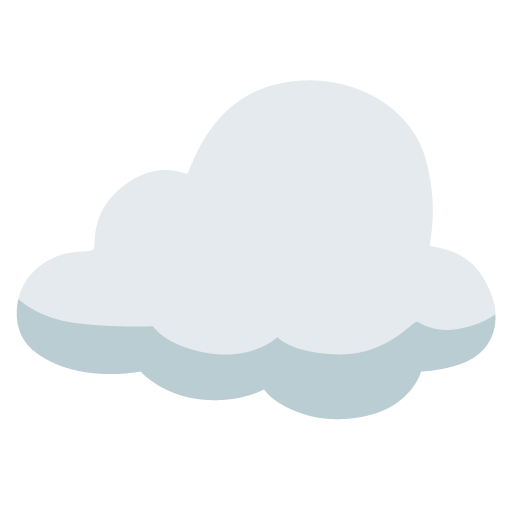} \includegraphics[width=1em]{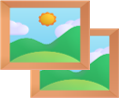}}"\xspace}
\newcommand{\secondcol}{``\raisebox{-0.2em}{\includegraphics[width=1em]{figures/elephant.png} \includegraphics[width=1em]{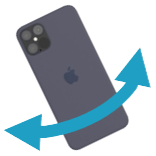}}"\xspace}
\newcommand{\thirdcol}{``\raisebox{-0.2em}{\includegraphics[width=1em]{figures/cloud.png} \includegraphics[width=1em]{figures/overlap.png}}"\xspace}
\newcommand{\fourthcol}{``\raisebox{-0.2em}{\includegraphics[width=1em]{figures/cam.png}}"\xspace}

\newcommand{\bgcolor}[2]{\setlength{\fboxsep}{0pt}\colorbox{#1}{\strut #2}}

\newcommand{\BGcolor}[3][HTML]{\definecolor{mycolor}{HTML}{#2}\bgcolor{mycolor}{#3}}

\newcommand{\davis}{\mbox{{DAVIS}}\xspace}
\newcommand{\dycheck}{\mbox{{DyCheck}}\xspace}
\newcommand{\adt}{\mbox{{ADT}}\xspace}
\newcommand{\tumd}{\mbox{{TUM-dynamics}}\xspace}
\newcommand{\JM}{{JM}}
\newcommand{\JR}{{JR}}

\usepackage{pifont}
\usepackage{bbding}
\newcommand{\cmark}{{\ding{51}}\xspace}
\newcommand{\xmark}{{\ding{55}}\xspace}

\usepackage{booktabs}
\usepackage{multirow}
\usepackage{tabularx}
\usepackage{tikz}

\definecolor{iccvblue}{rgb}{0.21,0.49,0.74}
\usepackage[pagebackref,breaklinks,colorlinks,allcolors=iccvblue]{hyperref}

\def\paperID{} %
\def\confName{ICCV}
\def\confYear{2025}

\usepackage[capitalize]{cleveref}

\newcommand{\REF}{\text{ref}}
\newcommand{\SRC}{\text{src}}
\newcommand{\DYN}{\text{dyn}}

\begin{document}

\title{\easier: Estimating Disentangled Motion from \duster Without Training}

\author{
Xingyu Chen$^{1,2}$ \quad Yue Chen$^{1,2}$ \quad Yuliang Xiu$^{2,3}$ \quad Andreas Geiger$^{4}$ \quad Anpei Chen$^{2,4}$ \vspace{2pt}\\
$^1$Zhejiang University \quad $^2$Westlake University 
\quad $^3$Max Planck Institute for Intelligent Systems \\
$^4$University of Tübingen, Tübingen AI Center \\
\href{https://easi3r.github.io/}{\texttt{\small easi3r.github.io}}
}

\maketitle

\begin{abstract}
Recent advances in \duster have enabled robust estimation of dense point clouds and camera parameters of static scenes, leveraging Transformer network architectures and direct supervision on large-scale 3D datasets.
In contrast, the limited scale and diversity of available 4D datasets present a major bottleneck for training a highly generalizable 4D model. This constraint has driven conventional 4D methods to fine-tune 3D models on scalable dynamic video data with additional geometric priors such as optical flow and depths. 
In this work, we take an opposite path and introduce \easier, a simple yet efficient training-free method for 4D reconstruction. Our approach applies attention adaptation during inference, eliminating the need for from-scratch pre-training or network fine-tuning. 
We find that the attention layers in \duster inherently encode rich information about camera and object motion. By carefully disentangling these attention maps, we achieve accurate dynamic region segmentation, camera pose estimation, and 4D dense point map reconstruction. 
Extensive experiments on real-world dynamic videos demonstrate that our lightweight attention adaptation significantly outperforms previous state-of-the-art methods that are trained or fine-tuned on extensive dynamic datasets.
\end{abstract}

\begin{figure}[t!]
    \centering
    \includegraphics[width=\linewidth,page=1]{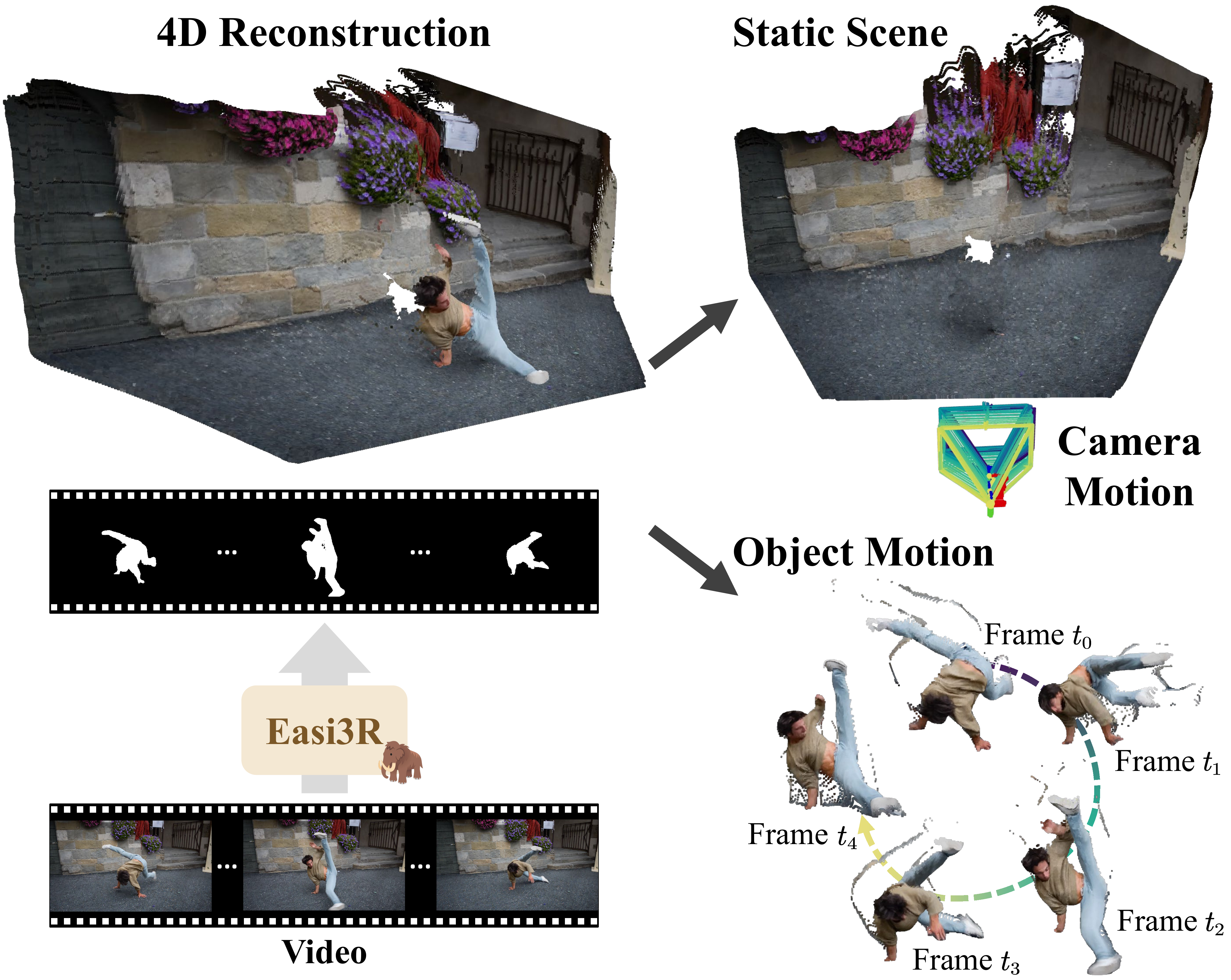}
    \caption{
        We present \easier, a training-free, plug-and-play approach that efficiently disentangles object and camera motion, enabling the adaptation of \duster for 4D reconstruction.
    }
    \label{fig:teaser}
\end{figure}

\section{Introduction}
Recovering geometry and motions from dynamic image collections is still a fundamental challenge in computer vision, with broad downstream applications in novel view synthesis, AR/VR, autonomous navigation, and robotics. 
The literature commonly identifies this problem as Structure-from-Motion (SfM) and has been the core focus in 3D vision over decades, yielding mature algorithms that perform well under stationary conditions and wide baselines.
However, these algorithms often fail when applied to dynamic video input.

The main reason for the accuracy and robustness gap between static and dynamic SfM is object dynamics, a common component in real-world videos. Moving objects violate fundamental assumptions of homography and epipolar consistency in traditional SfM methods~\cite{Schoenberger2016CVPR,pan2024glomap}.
In addition, in dynamic videos, where camera and object motions are often entangled, these methods struggle to disentangle the two motions, often causing the motion with rich texture to mainly contribute to camera pose estimation erroneously.
Recent efforts, such as \monster~\cite{monst3r} and \cuter~\cite{cut3r}, have made strides to address these challenges. 
However, their success is based on extensive training data~\cite{monst3r,MegaSaM,stereo4D,cut3r,das3r} or task-specific prior models~\cite{monst3r,MegaSaM,Robust-CVD,CasualSAM}, such as the depth, optical flow, and object mask estimators.
These limitations motivate us to innovate further to minimize the gap between static and dynamic reconstruction.

We ask ourselves if there are lessons from human perception that can be used as design principles for dynamic 4D reconstruction:
Human beings are capable of perceiving body motion and the structure of the scene, identifying dynamic objects, and disentangling ego-motion from object motion through the inherent attention mechanisms of the brain~\cite{Attentional_modulation}. 
Yet, the learning process rarely relies on explicit dynamic labels.

We observe that \duster implicitly learned a similar mechanism, and based on this, we introduce \easier, a training-free method to achieve dynamic object segmentation, dense point map reconstruction, and robust camera pose estimation from dynamic videos, as shown in \figref{fig:teaser}. 
\duster uses attention layers at its core, taking two image features as input and producing pixel-aligned point maps as output. These attention layers are trained to directly predict pointmaps in the reference view coordinate space, implicitly matching the image features between the input views~\cite{zeroCo} and estimating the rigid view transformation in the feature space.
In practice, performance drops significantly when processing pairs with object dynamics~\cite{monst3r}, as shown in \figref{fig:graph}.
By analyzing the attention maps in the transformer layers, we find that regions with less texture, under-observed, and dynamic objects can yield low attention values. Therefore, we propose a simple yet effective decomposition strategy to isolate the above components, which enables long-horizon dynamic object detection and segmentation.
With this segmentation, we perform a second inference pass by applying a re-weighting~\cite{prompt-to-prompt} in the cross-attention layers, enabling robust dynamic 4D reconstruction and camera motion recovery without fine-tuning on a dynamic dataset, all at minimal additional cost to \duster.

Despite its simplicity, we demonstrate that our inference-time scaling approach for 4D reconstruction is remarkably robust and accurate on in-the-wild casual dynamic videos. We evaluate our \easier adaptation on the \duster and \monster backbones in three task categories: camera pose estimation, dynamic object segmentation, and pointcloud reconstruction in dynamic scenes.
\easier performs surprisingly well across a wide range of datasets, even surpassing concurrent methods (\eg, \cuter~\cite{cut3r}, \monster~\cite{monst3r}, and \daser~\cite{das3r}) 
 that are trained on dynamic datasets.

\section{Related Work}

\boldparagraph{SfM and SLAM.}
Structure-from-Motion (SfM)~\cite{pollefeys1999self,pollefeys2004visual,snavely2008modeling,agarwal2011building,Schoenberger2016CVPR,snavely2006photo} and Simultaneous Localization and Mapping (SLAM)~\cite{newcombe2011dtam,mur2015orb,davison2007monoslam,engel2014lsd} have long been the foundation for 3D structure and camera pose estimation. These methods are done by associating 2D correspondences~\cite{bay2008speeded,mur2015orb,lowe2004distinctive,detone2018superpoint,sarlin2020superglue} or minimizing photometric errors~\cite{engel2014lsd,engel2017direct}, followed by bundle adjustment (BA)~\cite{agarwal2010bundle,triggs2000bundle,Droid,tang2018ba,Vggsfm,acezero} to refine structure and motion estimates. 
Although highly effective with dense input, these approaches often struggle with limited camera parallax or ill-posed conditions, leading to performance degeneracy.
To overcome these limitations, \duster~\cite{dust3r} introduced a learning-based approach that directly predicts two pointmaps from an image pair in the coordinate space of the first view. This approach inherently matches image features and rigid body view transformation. By leveraging a Transformer-based architecture~\cite{vit} and direct point supervision on large-scale 3D datasets, \duster establishes a robust Multi-View Stereo (MVS) foundation model.  
However, \duster and the follow-up methods~\cite{spann3r,DUSt3Rpp,SLAM3R,MASt3R-SLAM} assume primarily static scenes, which can lead to significant performance degradation when dealing with videos with dynamic objects~\cite{zhao2023pseudo}.

\boldparagraph{Pose-free Dynamic Scene Reconstruction.} 
Modifications to SLAM for dynamic scenes involve robust pose estimation to mitigate moving object interference, dynamic map management for updating changing environments, including techniques like semantic segmentation~\cite{DS-SLAM}, optical flows~\cite{zhao2022particlesfm}, enhance SLAM's resilience in dynamic scenarios.
Another line of work focuses on estimating stable video depth by incorporating geometric constraints~\cite{cvd}
and generative priors~\cite{hu2024depthcrafter, shao2024learning}. These methods enhance monocular depth accuracy but lack global point cloud lifting due to missing camera intrinsics and poses.
For joint pose and depth estimation, optimization-based methods such as CasualSAM~\cite{CasualSAM} fine-tune a depth network~\cite{MiDaS} at test time using pre-computed optical flow~\cite{RAFT}. Robust-CVD~\cite{Robust-CVD} refines pre-computed depth~\cite{MiDaS} and camera pose by leveraging masked optical flow~\cite{MaskRCNN, RAFT} to improve stability in occluded and moving regions. 
Concurrently, MegaSaM~\cite{MegaSaM} further enhances pose and depth accuracy by integrating DROID-SLAM~\cite{Droid}, optical flow~\cite{RAFT}, and depth initializations from~\cite{DepthAnything, UniDepth}, achieving state-of-the-art results.
Alternatively, point-map-based approaches like \monster~\cite{monst3r} extend \duster to dynamic scenes by fine-tuning with dynamic datasets and incorporating optical flow~\cite{RAFT} to infer dynamic object segmentation. \daser trains a DPT~\cite{dpt} on top of \monster, enabling feedforward segmentation estimation. \cuter~\cite{cut3r} fine-tunes \master~\cite{mast3r} on both static and dynamic datasets, achieving feedforward reconstruction but without predicting dynamic object segmentation, thereby entangling the static scene with dynamic objects.
Although effective, these methods require costly training on diverse motion patterns to generalize well.

In contrast, we take an opposite path, exploring a training-free and plug-in-play adaptation that enhances the generalization of \duster variants for dynamic scene reconstruction. Our method requires no fine-tuning and comes at almost no additional cost, offering a scalable and efficient alternative for handling real-world dynamic videos.

\boldparagraph{Motion Segmentation.}
Motion segmentation aims to predict dynamic object masks from video inputs. Classical approaches generally rely on optical flow estimation~\cite{yang2021self, xie2022segmenting, meunier2022driven, lian2023bootstrapping} and point tracking~\cite{ochs2013segmentation, sheikh2009background, yan2006general, brox2010object, karazija2025learning} to distinguish moving objects from the background. Being trained solely on 2D data, they often struggle with occlusions and distinguishing between object and camera motion.
To improve robustness, RoMo~\cite{goli2024romo} incorporates epipolar geometry~\cite{luong1996fundamental} and SAM2~\cite{sam2} to better disambiguate object motion from camera motion. While RoMo successfully combines COLMAP~\cite{Schoenberger2016CVPR} for accurate camera calibration, it focuses primarily on removing dynamic objects and reconstructing static scene elements only.
For the complete 4D reconstruction, \monster~\cite{monst3r} integrates optical flow~\cite{RAFT} with estimated pose and depth to predict dynamic object segmentation. \daser builds on \monster and trains a DPT~\cite{dpt} for segmentation inference. Using this segmentation, they align static components globally while preserving dynamic point clouds from each frame, enabling temporally consistent reconstruction of moving objects.

In this work, we discover that dynamic segmentation can be extracted from pre-trained 3D reconstruction models like \duster. We propose a simple, yet robust strategy to isolate this information from the attention layers, without the need for optical flow or pre-training on segmentation datasets.

\section{Method}
Given a casually captured video sequence $\{ I^t \in \mathbb{R}^{W\times H \times 3}\}_{t=1}^T$,
our goal is to estimate object and camera movements, as well as the canonical point clouds present in the input video. Object motion is represented as the segmentation sequence $\bM^t$, camera motion as the extrinsic and intrinsic pose sequences $\bP^t$, $\bK^t$, and point clouds $\cX^*$.
First, we formulate how the \duster model handles videos (\secref{ssec:video}). Next, we explore the mechanisms of attention aggregation in spatial and temporal dimensions (\secref{ssec:secrets}).
Finally, we introduce how aggregated cross-attention maps can be leveraged to decompose dynamic object segmentation (\secref{ssec:dynamic}), which in turn helps re-weight attention values for robust point cloud and camera pose reconstruction (\secref{ssec:reconstruction}).

\subsection{\duster with Dynamic Video}
\label{ssec:video}
\duster is designed for pose-free reconstruction, taking two RGB images -  $I^a, I^b$, where $a,b \in [1,\dots, T]$ - as input and output two pointmaps in the \emph{reference} view coordinate space, $X^{a \rightarrow a}, X^{b \rightarrow a} \in \mathbb{R}^{W\times H \times 3}$:
\begin{equation}
   X^{a \rightarrow a}, X^{b \rightarrow a} = \text{\duster}(I^a, I^b)
\end{equation}
Here, $X^{b \rightarrow a}$ denotes the pointmap of input $I^b$ predicted in the view $a$ coordinate space. 
In particular, both pointmaps are expressed in the \emph{reference} view coordinate, \ie, view $a$ in this example.

\begin{figure}[t!]
    \centering
    \includegraphics[width=1\linewidth,page=1]{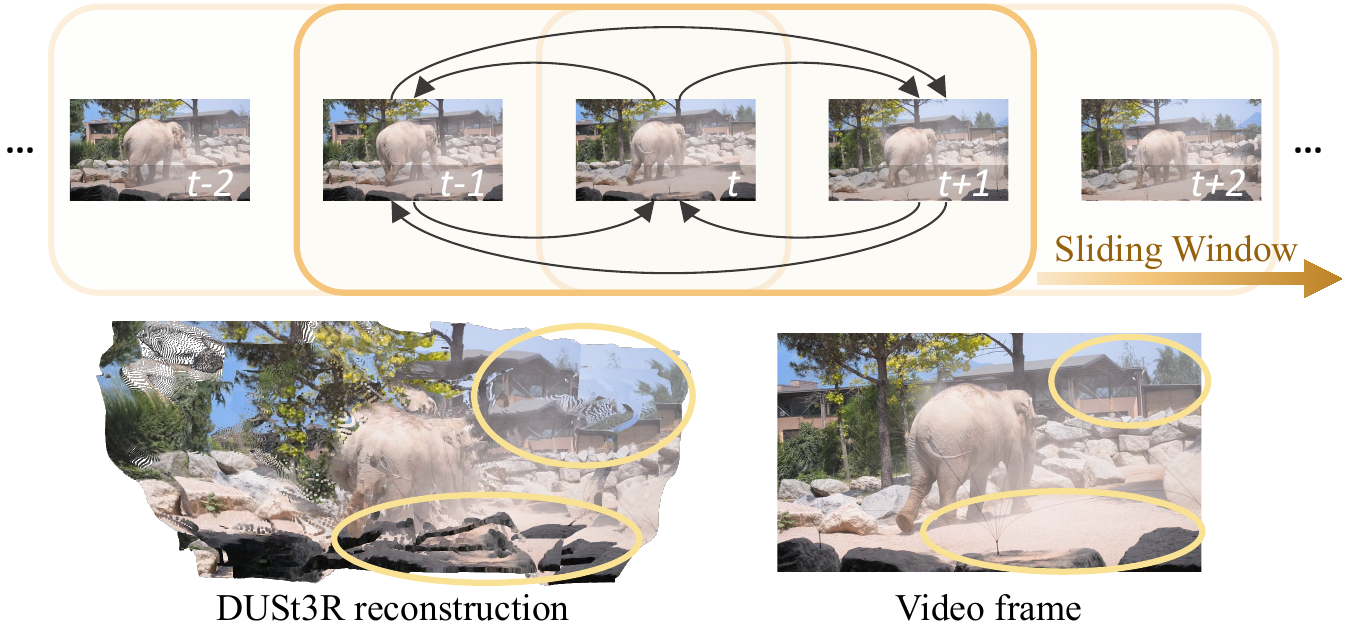}
    \caption{{\bf \duster with Dynamic Video.}
    We process videos using a sliding window and infer the \duster network pairwise. Reconstruction degrades with misalignment when dynamic objects occupy a considerable portion of the frames.
    }
    \label{fig:graph}
  \end{figure}

Given multi-view images, \duster processes them in pairs and globally aligns the pairwise predictions into a joint coordinate space using a connectivity graph across all views. However, this approach introduces computational redundancy for video sequences, as the view connectivity is largely known.
Instead, we process videos using a sliding temporal window
and infer the network for pair set $\varepsilon^{t} = \{(a, b) \mid a, b \in [t-\frac{n-1}{2},\dots,t+\frac{n-1}{2}], a \neq b \}$ within the symmetric temporal window of size $n$ centered at time $t$, as illustrated in the top row of \figref{fig:graph}. 

With pairwise predictions, it recovers globally aligned pointmaps $\{\cX^t  \in \mathbb{R}^{W\times H \times 3}\}_{t=1}^{T}$ by optimizing the transformation $\bP^t_i \in \mathbb{R}^{3 \times 4}$ from the coordinate space of each pair to the world coordinate, and a scale factor $\bs^t_i$ for the $i$-th pair within the set of pairs $\varepsilon^{t}$:

\begin{multline}
\cX^* = \argmin_{\cX,\bP,\bs} \sum_{t \in T} \sum_{ i \in \varepsilon^t} \|\cX^a - \bs^t_i \bP^t_i X^{a \rightarrow a}\|_1 \\
  + \| \cX^b - \bs^t_i \bP^t_i X^{b \rightarrow a}\|_1
\label{eq:recon_objective}
\end{multline}

Note that the above optimization process assumes a reliable pairwise reconstruction and that global content can be registered by minimizing the linear equations in \eqnref{eq:recon_objective}.
However, since \duster are learned from RGB-D images of static scenes, dynamic objects disrupt the learned epipolar matching policy. 
As a result, registration fails when dynamic content occupies a considerable portion of pixels, as shown in \figref{fig:graph}.

\begin{figure*}[t!]
    \centering
    \includegraphics[width=1\linewidth,page=1]{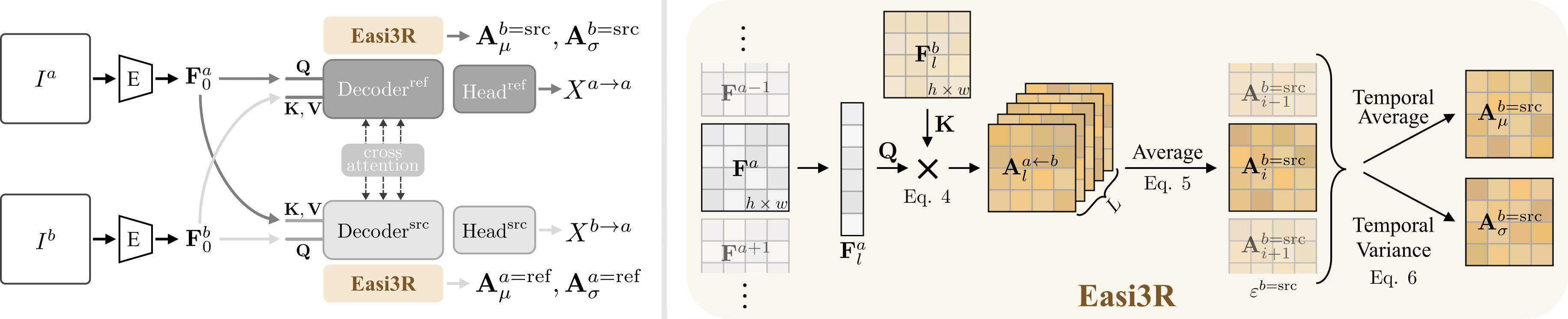}
    \caption{{\bf \duster } and our {\bf \easier} adaptation. \duster encodes two images $I^a,I^b$ into feature tokens $\bF_0^a,\bF_0^b$, which are then decoded into point maps in the reference view coordinate space using two decoders. Our \easier aggregates the cross-attention maps from the decoders, producing four semantically meaningful maps: $\bA^{b=\SRC}_\mu,\bA^{b=\SRC}_\sigma,\bA^{a=\REF}_\mu,\bA^{a=\REF}_\sigma$. These maps are then used for a second inference pass to enhance reconstruction quality.}
    \label{fig:pipe}
  \end{figure*}

\subsection{Secrets Behind \duster}
\label{ssec:secrets}

We now examine the network architecture to identify the components that cause failures for dynamic video input.
As shown in \figref{fig:pipe}, \duster consists of two branches: the top one for the reference image $I^a$ and the bottom one for the source image $I^b$.
The two input images are first processed by a weight-sharing ViT encoder~\cite{vit}, producing token representations $\bF_0^a = \text{Encoder}(I^a)$ and $\bF_0^b = \text{Encoder}(I^b)$.
Next, two decoders, composed of a sequence of decoder blocks, exchange information both within and between views. In each block, self-attention is applied to the token outputs from the previous block, while cross-attention is performed using the corresponding block outputs from the other branch.

\begin{equation}
  \begin{aligned}
    \bF_l^a &= \text{DecoderBlock}^\REF_{l}\left(\bF_{l-1}^a, \bF_{l-1}^b\right) \\
    \bF_l^b &= \text{DecoderBlock}^\SRC_{l}\left(\bF_{l-1}^b, \bF_{l-1}^a\right)
  \end{aligned}
  \label{eq:decoder}
\end{equation}
where $l = 1,\ldots,L$ is the block index. 
Using the feature tokens, two regression heads produce pointmap predictions: $X^{a \rightarrow a} = \text{Head}^\REF\left(\bF_0^a, \dots, \bF_L^a\right)$ and $X^{b \rightarrow a} = \text{Head}^\SRC\left(\bF_0^b, \dots, \bF_L^b\right)$, respectively.
The blocks are trained by minimizing the Euclidean distance between the predicted and ground-truth pointmaps.

\begin{figure}[t!]
    \centering
    \includegraphics[width=1\linewidth,page=1]{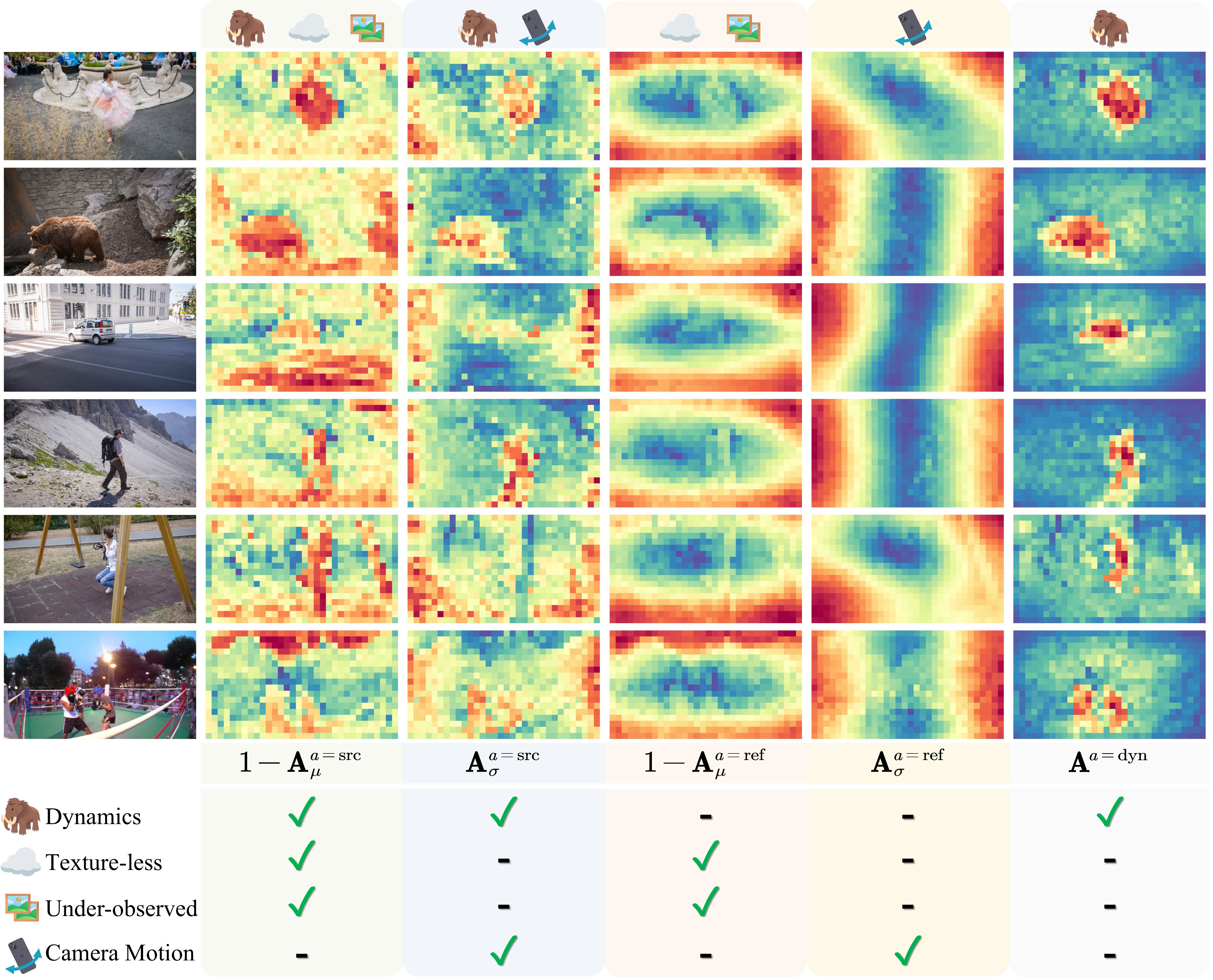}
    \caption{{\bf Visualization for Cross-Attention Maps.} We color the \emph{normalized} values of attention maps, ranging from \BGcolor{f06744}{o}\BGcolor{f88d51}{n}\BGcolor{fdb466}{e}\BGcolor{fdd380}{ }\BGcolor{feeb9e}{t}\BGcolor{ffffbf}{o}\BGcolor{eff8a6}{ }\BGcolor{d7ef9b}{z}\BGcolor{b2e0a2}{e}\BGcolor{88cfa4}{r}\BGcolor{5fbaa8}{o}. We highlight the patterns captured by each type of attention map using relatively high values. For a more detailed demonstration, we invite reviewers to visit our webpage under \href{https://easi3r.github.io/}{easi3r.github.io}.}
    \label{fig:attn_vis}
  \end{figure}

\boldparagraph{Observation.}
Our key insight is that \duster implicitly learns rigid view transformations through its cross-attention layers, assigning low attention values to tokens that violate epipolar geometry constraints, such as texture-less, under-observed, and dynamic regions.  
By aggregating cross-attention outputs across spatial and temporal dimensions, we extract motions from the attention layers.

\boldparagraph{Spatial attention maps.}
As illustrated in the \figref{fig:pipe} (left), the image features $\bF$ are projected into a query matrix for their respective branch with $\bQ = \ell_\bQ(\bF) \in \mathbb{R}^{(h \times w) \times c}$, while also serving as a key and value matrix for the other matrices, $\bK = \ell_\bK(\bF)  \in \mathbb{R}^{(h \times w) \times c}$, where $c$ is the feature dimenson. The projections are obtained using trainable linear functions $\ell_\bQ(\cdot), \ell_\bK(\cdot)$.
As illumistracted in the right side of~\figref{fig:pipe}, this results in the cross-attention map:
\begin{equation}
\bA_l^{a \leftarrow b}=\bQ_l^a{\bK_l^b}^T/\sqrt{c}, \quad \bA_l^{b \leftarrow a}=\bQ_l^b{\bK_l^a}^T/\sqrt{c}
\end{equation}
in which the cross-attention map $\bA_l^{a \leftarrow b}, \quad \bA_l^{b \leftarrow a} \in \mathbb{R}^{(h \times w) \times h \times w}$ are used to guide the warping of the value matrix $\bV = \ell_\bV(\bF) \in \mathbb{R}^{(h \times w) \times c}$ , and the cross-attention output in the reference view branch is given by $\mathrm{softmax}(\bA_l^{a \leftarrow b})\bV^b$. Intuitively, the attention map $\bA_l^{a \leftarrow b}$ determines how the information is aggregated from the view $b$ to the view $a$ in the $l$-th decoder block.

To evaluate the spatial contribution of each token in view $b$ to all tokens in view $a$, we average the attention values between different tokens along the query and layer dimensions. This is given by,

\begin{equation}
  \begin{aligned}
    \bA^{b=\SRC} &= \sum_l \sum_x \bA_l^{a \leftarrow b}(x,y,z) / (L \times h \times w) \\
    \bA^{a=\REF} &= \sum_l \sum_x \bA_l^{b \leftarrow a}(x,y,z) / (L \times h \times w)
  \end{aligned}
\end{equation}
where $\bA^{b=\SRC}, \bA^{a=\REF} \in \mathbb{R}^{h \times w}$, representing the averaged attention maps, capturing the overall influence of tokens from one view to another across all decoder layers. \ie, $\bA^{b=\SRC}$ denotes the overall contribution of view $b$ to the reference view when it serve as source view.

\boldparagraph{Temporal attention maps.}
In the following, we extend the above single-pair formulation to multiple pairs to explore their temporal attention correlations.
For a specific frame $I^t$, it pairs with multiple frames, resulting in $2(n-1)$ attention maps per frame. As shown in the upper row of \figref{fig:graph}, window size of 3 corresponds to 4 pairs.
To aggregate the pairwise cross-attention maps temporally, we compute the mean and variance over pairs that the view serves as source and reference: 
\begin{equation}
\bA^{b=\SRC}_{\mu}  = \textrm{Mean}(\bA_i^{b=\SRC}), \,\,
\bA^{b=\SRC}_{\sigma} = \textrm{Std}(\bA_i^{b=\SRC})
\end{equation}
where $i \in \varepsilon^{b=\SRC}$ and
\begin{equation}
\varepsilon^{b=\SRC}=\{(a,b) \mid \SRC=b, a \in [t-n,\dots,t+n], a \neq b \}
\end{equation}

Similarly, we compute $\bA^{b=\REF}_{\mu}$ and $\bA^{b=\REF}_{\sigma}$ for the set of pairs where view $a$ acts as the source view:  
\begin{equation}
\varepsilon^{b=\REF}=\{(b,a) \mid \REF=b, a \in [t-n,\dots,t+n], a \neq b \}
\end{equation}

\boldparagraph{Secrets.}
We visualize the aggregated temporal cross-attention maps in~\figref{fig:attn_vis}. Recall that \duster infers pointmaps from two images in the reference view coordinate frame, implicitly aligning points from the source view to the reference view.

(\rmnum{1}) The reference view serves as the registration standard and is assumed to be static. As a result, the average attention map $\bA^{a=\REF}_{\mu}$ tends to be smooth, with texture-less regions (\eg, ground, sky, swing supports, boxing fences) and under-observed areas (\eg, image boundary) naturally exhibiting low attention values, since \duster believes that they are less useful for registration. These regions can be highlighted and extracted using $(1-\bA_{\mu}^{a=\REF})$, as shown in the \thirdcol column of~\figref{fig:attn_vis}.

(\rmnum{2}) By calculating the standard deviation of $(1-\bA_{\mu}^{a=\REF})$ between neighboring frames, we have $\bA_{\sigma}^{a=\REF}$, \eg, the column \fourthcol, representing the changes of the token contribution in the image coordinate space. Pixels perpendicular to the direction of motion generally share similar pixel flow speeds, resulting in consistent deviations that allow us to infer camera motion from the attention pattern. For example, in the``Walking Man" case in the fourth row of the~\figref{fig:attn_vis}, with the camera motion from left to right, we can observe pixels along a column sharing similar attention values.

(\rmnum{3}) Similar to the reference view, we also compute the average invert attention map in the source view, $1 - \bA^{a=\SRC}_{\mu}$. As shown in the \firstcol column, the result not only indicates areas with less texture and underobserved areas but also highlights dynamic objects because they violate the rigid body transformation prior which \duster has learned from the 3D dataset, resulting in low $\bA^{a=\SRC}_{\mu}$ values.

(\rmnum{4}) The column \secondcol shows the standard deviation of the source view attention map, $\bA_{\sigma}^{a=\SRC}$. It highlights both camera and object motion, as the attention of these areas continuously changes over time, leading to high deviation in image space.

\subsection{Dynamic Object Segmentation}
\label{ssec:dynamic}

By observing the compositional properties of the derived cross-attention maps, we propose extracting dynamic object segmentation for free, which provides a key for bridging static and dynamic scene reconstruction.
To this end, we identify attention activations attributed to object motion. We infer the dynamic attention map for frame $a$ by computing the joint attention of the first two attention columns in \figref{fig:attn_vis} using the element-wise product: $(1-\bA^{a=\SRC}_{\mu}) \cdot \bA^{a=\SRC}_{\sigma}$.
To further mitigate the effects of texture-less regions, under-observed areas, and camera motion (as shown in the third and fourth columns of \figref{fig:attn_vis}), we incorporate the outputs with their inverse attention, resulting in the final formula: 
\begin{equation}
\bA^{a=\DYN} = (1-\bA^{a=\SRC}_{\mu}) \cdot \bA^{a=\SRC}_{\sigma} \cdot \bA^{a=\REF}_{\mu} \cdot (1-\bA^{a=\REF}_{\sigma})
\label{eq:dynamic_seg}
\end{equation}
we then obtain per-frame dynamic object segmentation $\bM^{t} = \left [\bA^{t=\DYN} > \alpha\right]$ using ~\eqnref{eq:dynamic_seg} and $\bM^{t}\in \mathbb{R}^{h \times w}$, $\alpha$ denotes a pre-defined attention threshold and the $[\cdot]$ is the Iverson bracket. Note that the segmentation is processed frame by frame. To enhance temporal consistency, we apply a feature clustering method that fuses information across all frames; see the supplementary materials for more details.

\subsection{4D Reconstruction}
With dynamic object segmentation, the most intuitive way to adapt static models to dynamic scenes is by masking out dynamic objects during inference at both the image and token levels. This can be done by replacing dynamic regions with black pixels in the image and substituting the corresponding tokens with mask tokens.
In practice, this approach significantly degrades reconstruction performance~\cite{monst3r}, mainly because black pixels and mask tokens lead to out-of-distribution input.  
This motivates us to apply masking directly within the attention layers instead of modifying the input images.

\label{ssec:reconstruction}
\boldparagraph{Attention re-weighting.}
We propose to modify the cross-attention maps by weakening the attention values associated with dynamic regions.
To achieve this, we perform a second inference pass through the network, masking the attention map for assigned dynamic regions. This results in zero attention for those regions while keeping the rest of the attention maps unchanged:
\begin{equation}
  \begin{aligned}
    \mathrm{softmax}(\tilde{\bA}_l^{a \leftarrow b}) = \begin{cases}
      0 & \text{if }  \bM^{a \leftarrow b} \\
      \mathrm{softmax}(\bA_l^{a \leftarrow b}) & \text{otherwise}
    \end{cases}
  \end{aligned}
\end{equation}
here, $\bM^{a \leftarrow b} = (1 - \bM^a) \otimes {\bM^b}^T$, where $\bM^{a \leftarrow b} \in \mathbb{R}^{(h \times w)\times (h \times w)}$ and $\otimes$ denotes outer product. This results in tokens from dynamic regions in view $b$ that do not contribute to static regions in view $a$.
It is important to note that re-weighting is applied only to the reference view decoder, as source view requires a static reference (i.e., the reference view), as described in the secret (\rmnum{1}).  
To achieve this, the source view decoder must perform cross-attention with all tokens from the reference view. Re-weighting dynamic attention on both branches could result in the loss of static standard, leading to noisy outputs.  
We conducted an ablation study on this insight.

\boldparagraph{Global alignment.}
We align the predicted pointmaps from the sliding windows with the global world coordinate using \eqnref{eq:recon_objective}.  
Moreover, thanks to dynamic region segmentation, our method also supports segmentation-aware global alignment with optical flow.
In particular, we incorporate a reprojection loss to ensure that the projected point flow remains consistent with the optical flow estimation~\cite{RAFT}.  
Specifically, given an image pair $(a,b)$, we compute the camera motion  from frame $a$ to frame $b$, denoted by $\hat{\mathcal{F}}^{a \rightarrow b}$, by projecting the global point map $\cX^b$ from camera $(\bP^a,\bK^a)$ to camera $(\bP^{b},\bK^b)$.  
We then enforce the consistency between the computed flow and the estimated optical flow $\mathcal{F}^{a \rightarrow b}$ in static regions $(1-\bM^{t})$:

\begin{multline}
  \mathcal{L}_{\textrm{flow}}  =  \sum_{t \in T} \sum_{ i \in \varepsilon^t} (1-\bM^a) \cdot \|\hat{\mathcal{F}}_i^{a \rightarrow b} - \mathcal{F}_i^{a \rightarrow b}\|_1 \\
  + (1-\bM^{b}) \cdot \|\hat{\mathcal{F}}_i^{b \rightarrow a} - \mathcal{F}_i^{b \rightarrow a}\|_1
\end{multline}
where $\cdot$ indicates element-wise product. By incorporating flow constraint into the optimization process in~\eqnref{eq:recon_objective}, we achieve a more robust output in terms of global pointmaps $\cX^*$ and pose sequences $\bP^t$, $\bK^t$. Note that this term is used optionally to ensure a fair comparison with the baseline that does not incorporate the flow-estimation model.

\section{Experiments}
\label{ssec:exp}
We evaluate our method in a variety of tasks, including dynamic object segmentation (\secref{ssec:exp_dyn_seg}), camera pose estimation (\secref{ssec:exp_pose_est}) and 4D reconstruction (\secref{ssec:exp_4drecon}).
We performed ablation studies in supplementary.

\boldparagraph{Baselines.}
We compare \easier with state-of-the-art pose free 4D reconstruction method, including \duster~\cite{dust3r}, \monster~\cite{monst3r}, \daser~\cite{das3r}, and \cuter~\cite{cut3r}. Among these works, the latter three are concurrent works that also aim to extend \duster to handle dynamic videos, but take a different approach by fine-tuning on dynamic datasets, such as~\cite{TartanAir,Waymo,DynamicReplica},
and optimization under the supervision of optical flow~\cite{RAFT}.
Unlike previous work, our method performs a second inference pass on top of the pre-trained \duster or \monster model without requiring fine-tuning or optimization on additional data.

\begin{figure}[t!]
    \centering
    \includegraphics[width=1\linewidth,page=1]{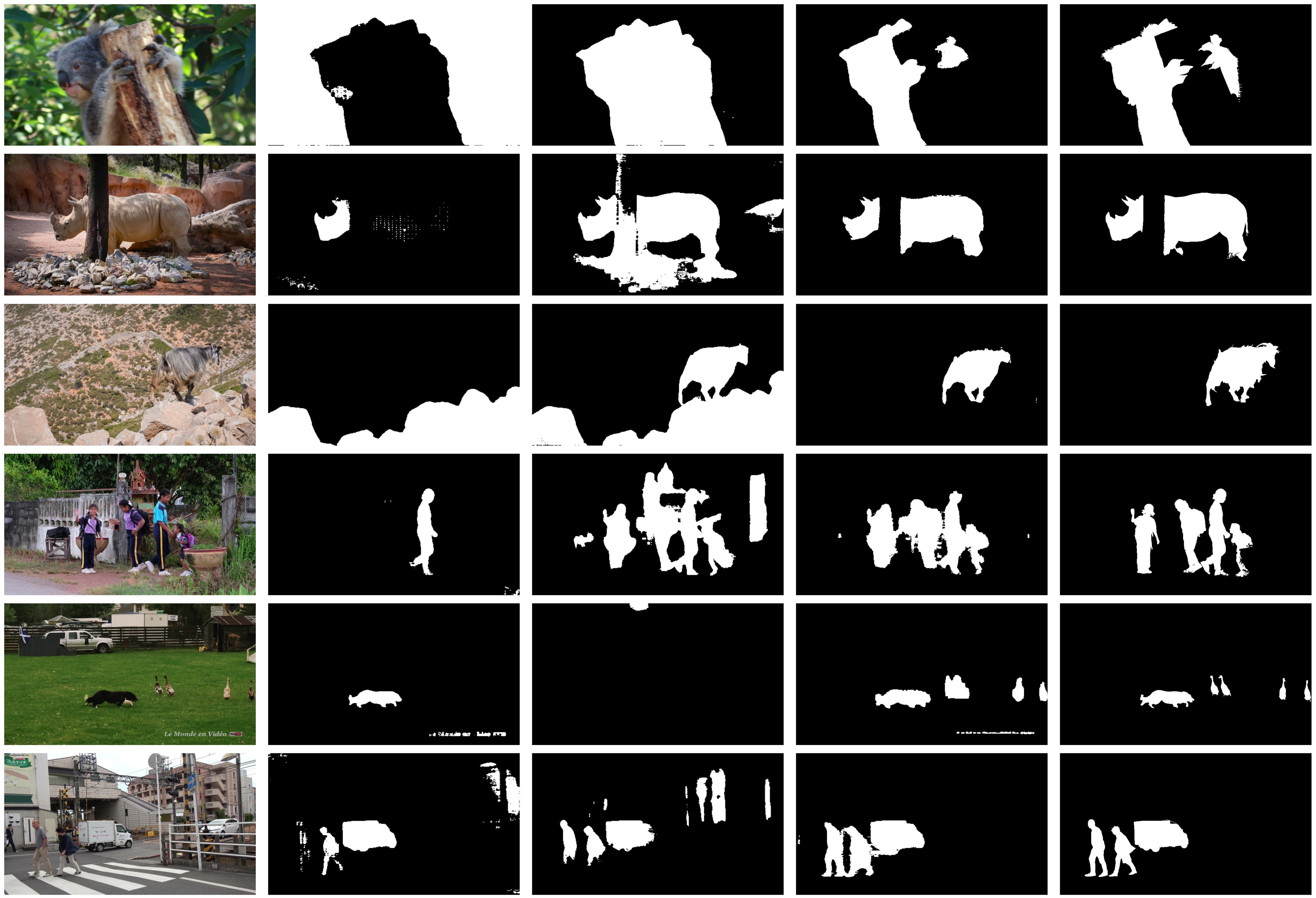}
     \setlength{\tabcolsep}{0.05pt}     %
       \scriptsize
          \begin{tabularx}{\linewidth}{
             >{\centering\arraybackslash}m{0.2\linewidth}
             >{\centering\arraybackslash}m{0.2\linewidth}
             >{\centering\arraybackslash}m{0.2\linewidth}
             >{\centering\arraybackslash}m{0.2\linewidth}
             >{\centering\arraybackslash}m{0.2\linewidth}
            }
            Video Frame & \monster~\cite{monst3r} & \daser~\cite{das3r} & Ours & GT 
             \end{tabularx}
    \caption{{\bf Qualitative Results of Dynamic Object Segmentation.}
    ``Ours" refers to the $ \easier_\text{monst3r}$ setting. Here, we present the enhanced setting, where outputs from different methods serve as prompts and are used with SAM2~\cite{sam2} for mask inference.
    }
    \label{fig:mask}
  \end{figure}

\begin{table}[t]
  \centering
  \scriptsize
  \renewcommand{\tabcolsep}{2pt}
  \caption{\textbf{Dynamic Object Segmentation} on the \davis dataset. The best and second best results are \textbf{bold} and \underline{underlined}, respectively. \easier$_\text{dust3r/monst3r}$ denotes the \easier experiment with the backbones of \monster/\duster.}
  \label{tab:exp_dyn_seg}
  \resizebox{1.0\linewidth}{!}{
  \begin{tabular}{lc|llll|llll|llll} 
  \toprule
   & & \multicolumn{4}{c|}{\davis-16} & \multicolumn{4}{c|}{\davis-17} & \multicolumn{4}{c}{\davis-all} \\ 
   
   \cmidrule(lr){3-6} \cmidrule(lr){7-10} \cmidrule(lr){11-14} 
   
   & & \multicolumn{2}{c}{w/o SAM2} & \multicolumn{2}{c|}{w/ SAM2} & \multicolumn{2}{c}{w/o SAM2} & \multicolumn{2}{c|}{w/ SAM2} & \multicolumn{2}{c}{w/o SAM2} & \multicolumn{2}{c}{w/ SAM2}\\ 
   
  Method & Flow & {\JM $\uparrow$} & {\JR $\uparrow$} & {\JM $\uparrow$} & {\JR $\uparrow$} & {\JM $\uparrow$} & {\JR $\uparrow$} & {\JM $\uparrow$} & {\JR $\uparrow$} & {\JM $\uparrow$} & {\JR $\uparrow$} & {\JM $\uparrow$} & {\JR $\uparrow$}\\ 
  \midrule
  
  \duster~\cite{dust3r} & \cmark & 42.1 & 45.7 & 58.5 & 63.4 & 35.2 & 35.3 & 48.7 & 50.2 & 35.9 & 34.0 & 47.6 & 48.7 \\
  
  \monster~\cite{monst3r} & \cmark & 40.9 & 42.2 & 64.3 & \underline{73.3} & 38.6 & 38.2 & 56.4 & 59.6 & 36.7 & 34.3 & 51.9 & 54.1 \\
  
  \daser~\cite{das3r} &\xmark & 41.6 & 39.0 & 54.2 & 55.8 & 43.5 & 42.1 & 57.4 & 61.3 & 43.4 & 38.7 & 53.9 & 54.8 \\
  
  \textbf{\easier}$_\text{dust3r}$ & \xmark & \underline{53.1} & \underline{60.4} & \underline{67.9} & 71.4 & \underline{49.0} & \underline{56.4} & \underline{60.1} & \underline{65.3} & \underline{44.5} & \underline{49.6} & \underline{54.7} & \underline{60.6} \\
  
  \textbf{\easier}$_\text{monst3r}$ & \xmark & \textbf{57.7} & \textbf{71.6} & \textbf{70.7} & \textbf{79.9} & \textbf{56.5} & \textbf{68.6} & \textbf{67.9} & \textbf{76.1} & \textbf{53.0} & \textbf{63.4} & \textbf{63.1} & \textbf{72.6} \\
  \bottomrule
  \end{tabular}
  }
\end{table}

\subsection{Dynamic Object Motion}
\label{ssec:exp_dyn_seg}
We represent object motion as a segmentation sequence and evaluate performance on the video object segmentation benchmark \davis-16~\cite{davis}, more challenging \davis-17~\cite{davis2017}, and \davis-all.
We present two experiment settings: direct evaluation of network outputs and an enhanced setting where outputs serve as prompts for SAM2~\cite{sam2}, improving results. These settings are denoted as w/ and w/o SAM2 in \tabref{tab:exp_dyn_seg}. 
Following \davis~\cite{davis}, we evaluate performance using IoU mean (JM) and IoU recall (JR) metrics.
Since \duster originally does not support dynamic object segmentation, we extend it as a baseline by incorporating the flow-guided segmentation as \monster. 
By applying our attention-guided decomposition, both \duster and \monster show improved segmentation, without the need for flow, even surpassing \daser, which is explicitly trained on dynamic mask labels.

\boldparagraph{Qualitative Results.} \figref{fig:mask} presents the qualitative comparison between our method and existing approaches.  
Since \monster relies on optical flow estimation, it struggles in textureless regions, failing to disentangle dynamic objects from the background (\eg, koala, rhino, sheep).  
On the other hand, \daser learns a mask head for dynamic segmentation but tends to over-segment in most cases.  
Our method, built on \duster and enhanced with our \easier attention-guided decomposition, accurately segments dynamic objects while maintaining robustness in handling textureless regions (\eg, trunks, rocks, walls), small dynamic objects (\eg, goose), and casual motions (\eg, girls, pedestrian).  
The results provide a surprising insight that 3D models, such as \duster in our case, may inherently possess a strong understanding of the scene and can generalize well to standard 2D tasks.

\begin{figure*}[t!]
    \centering
      \small
     \renewcommand{\arraystretch}{0.5} %
     \setlength{\tabcolsep}{0.005pt}     %
  \begin{tabularx}{\linewidth}{
     >{\centering\arraybackslash}m{0.05\linewidth}
     >{\centering\arraybackslash}m{0.2375\linewidth}
     >{\centering\arraybackslash}m{0.2375\linewidth}
     >{\centering\arraybackslash}m{0.2375\linewidth}
     >{\centering\arraybackslash}m{0.2375\linewidth}
    }
    Video & \quad \cuter~\cite{cut3r} & \monster~\cite{monst3r} & \daser~\cite{das3r} & Ours 
     \end{tabularx}
    \includegraphics[width=1.0\linewidth,page=1]{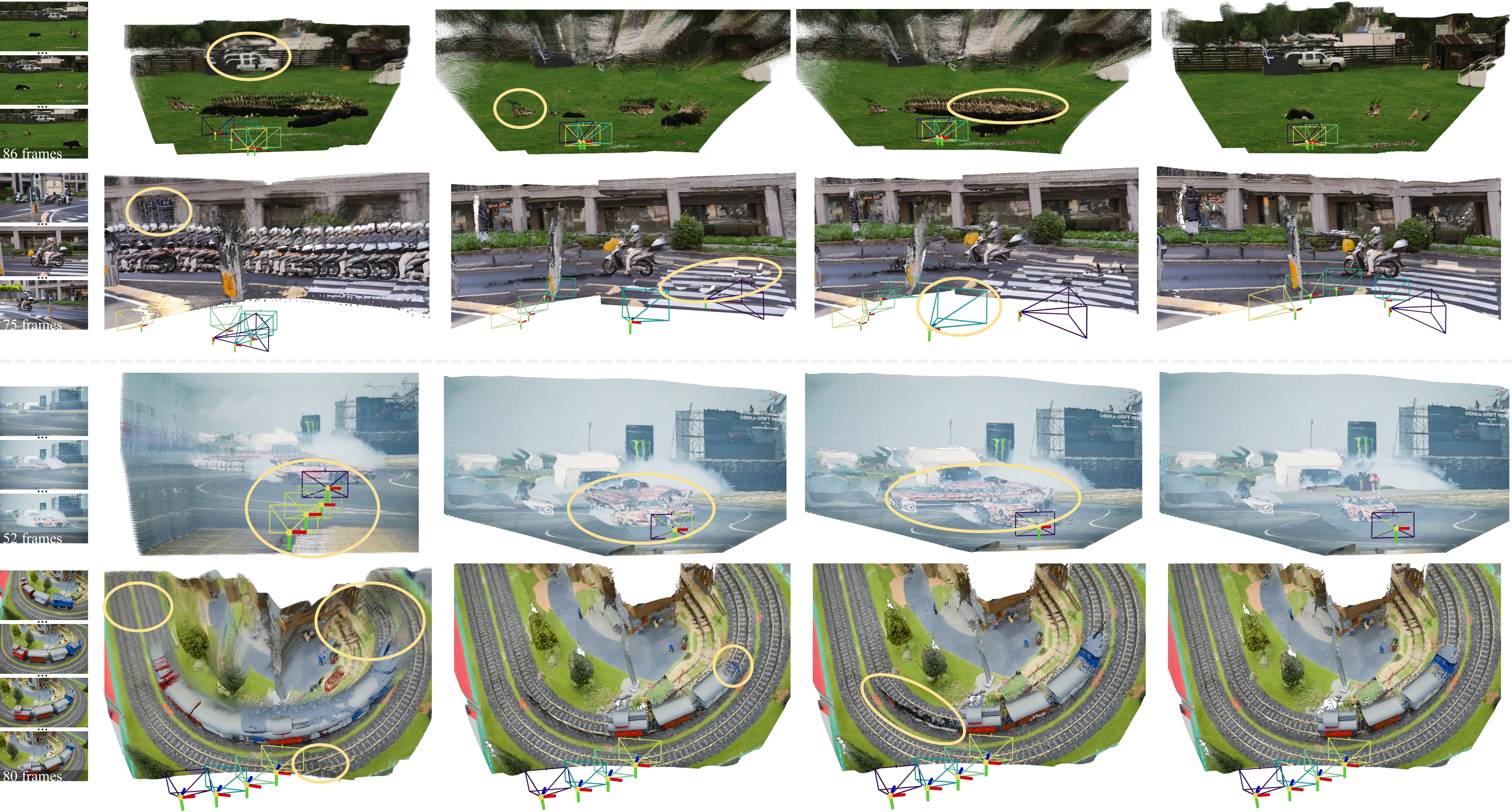}
    \caption{{\bf Qualitative Comparison.} 
    We visualize cross-frame globally aligned static scenes with dynamic point clouds at a selected timestamp. Notably, instead of using ground truth dynamic masks in previous work, we apply the estimated per-frame dynamic masks to filter out dynamic points at other timestamps for comparison.
    Our method (top two and bottom two rows as \easier$_\text{dust3r/monst3r}$, respectively) achieves temporally consistent reconstruction of both static scenes and moving objects, whereas baselines suffer from static structure misalignment and unstable camera pose estimation, and ghosting artifacts due to inaccuracy estimation of dynamic segmentation.}
    \label{fig:vis_pts}
  \end{figure*}

\subsection{Camera Motion}
\label{ssec:exp_pose_est}
We evaluate camera motion by using the estimated extrinsic sequence on three dynamic benchmarks: \dycheck~\cite{dycheck}, \tumd~\cite{tumd}, and \adt~\cite{adt,tapvid} datasets. 
Specifically, the \adt dataset features egocentric videos, which are out-of-distribution for \duster's training set. The \dycheck dataset includes diverse, in-the-wild dynamic videos captured from handheld cameras. The \tumd dataset contains major dynamic objects in relatively simple indoor scenarios.
Instead of evaluating video clips as in previous methods, we adopt a more challenging setting by processing \textbf{entire sequences}. Specifically, we downsample frames at different rates: every 5 frames for \adt, 10 for \dycheck, and 30 for \tumd, resulting in approximately 40 frames.
We use standard error metrics: Absolute Translation Error (ATE), Relative Translation Error (RTE), and Relative Rotation Error (RRE), after applying the Sim(3) alignment~\cite{umeyama} on the estimated camera trajectory to the GT.

\begin{table}[t]
  \centering
  \scriptsize
  \renewcommand{\tabcolsep}{2pt}
  \caption{\textbf{Benefits of Easi3R on Camera Pose Estimation} on the DyCheck, ADT and \tumd datasets. The best and second best results are \textbf{bold} and \underline{underlined}, respectively. \easier$_\text{dust3r/monst3r}$ denotes the \easier experiment with the backbones of \monster/\duster.}
  \label{tab:camera_pose_benefit}
  \resizebox{\linewidth}{!}{
  \begin{tabular}{@{}lc|lll|lll|lll@{}}
  \toprule
  & & \multicolumn{3}{c}{DyCheck} & \multicolumn{3}{c}{ADT} & \multicolumn{3}{c}{\tumd} \\ 
  \cmidrule(lr){3-5} \cmidrule(lr){6-8} \cmidrule(lr){9-11}
  {Method}  & Flow & {ATE $\downarrow$} & {RTE $\downarrow$} & {RRE $\downarrow$} & 
  {ATE $\downarrow$} & {RTE $\downarrow$} & {RRE $\downarrow$} & 
  {ATE $\downarrow$} & {RTE $\downarrow$} & {RRE $\downarrow$} \\ 
  \midrule
  
  \duster~\cite{dust3r} & \xmark & 0.035 & 0.030 & 2.323 & \underline{0.042} & 0.025 & 1.212 & 0.100 & 0.087 & 2.692 \\

  \textbf{\easier}$_\text{dust3r}$ & \xmark & \underline{0.029} & 0.025 & \underline{1.774} & \textbf{0.040} & \underline{0.021} & \underline{0.880} & 0.093 & 0.076 & \underline{2.366} \\

  \duster~\cite{dust3r} & \cmark & \underline{0.029} & \underline{0.021} & 1.875 & 0.076 & 0.030 & 0.974 & \underline{0.071} & \underline{0.067} & 3.711 \\
  
  \textbf{\easier}$_\text{dust3r}$ & \cmark & \textbf{0.021} & \textbf{0.014} & \textbf{1.092} & \underline{0.042} & \textbf{0.015} & \textbf{0.655} & \textbf{0.070} & \textbf{0.061} & \textbf{2.361} \\
  
  \midrule
  
  \monster~\cite{monst3r} & \xmark & 0.040 & 0.034 & 1.820 & \underline{0.045} & \underline{0.024} & 0.759 & 0.183 & \textbf{0.148} & 6.985 \\

  \textbf{\easier}$_\text{monst3r}$ & \xmark & 0.038 & 0.032 & 1.736 & \underline{0.045} & \underline{0.024} & \underline{0.715} & 0.184 & \underline{0.149} & \underline{6.311} \\

  \monster~\cite{monst3r} & \cmark & \underline{0.033} & \underline{0.024} & \underline{1.501} & 0.055 & 0.025 & 0.776 & \underline{0.170} & 0.155 & 6.455 \\

  \textbf{\easier}$_\text{monst3r}$ & \cmark & \textbf{0.030} & \textbf{0.021} & \textbf{1.390} & \textbf{0.039} & \textbf{0.016} & \textbf{0.640} & \textbf{0.168} & 0.150 & \textbf{5.925} \\
  
  \bottomrule
  \end{tabular}
    }
\end{table}

\begin{figure}[t!]
  \centering
  \begin{minipage}[b]{0.445\linewidth}
    \centering
    \includegraphics[width=\linewidth]{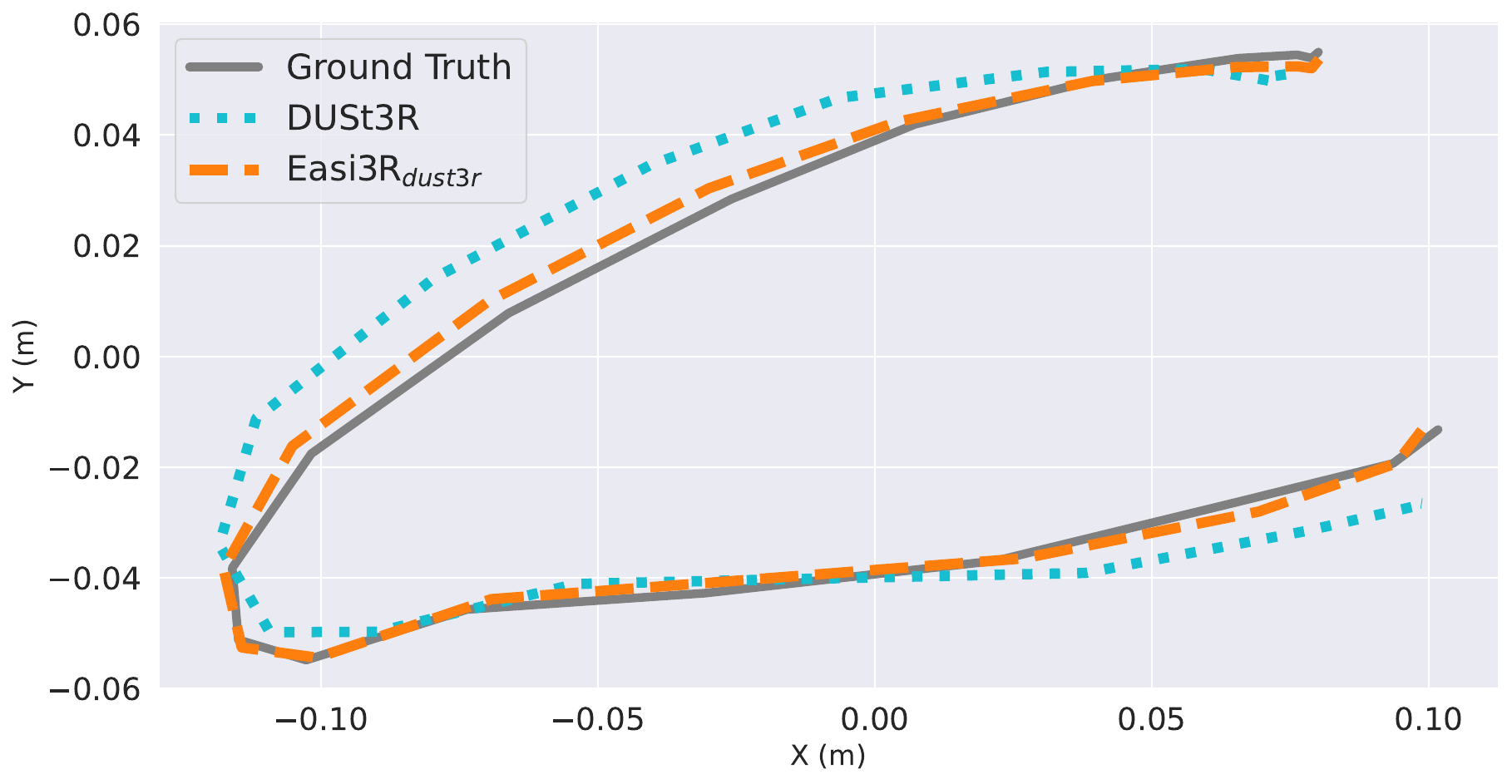}
  \end{minipage}%
  \begin{minipage}[b]{0.55\linewidth}
    \centering
    \includegraphics[width=\linewidth]{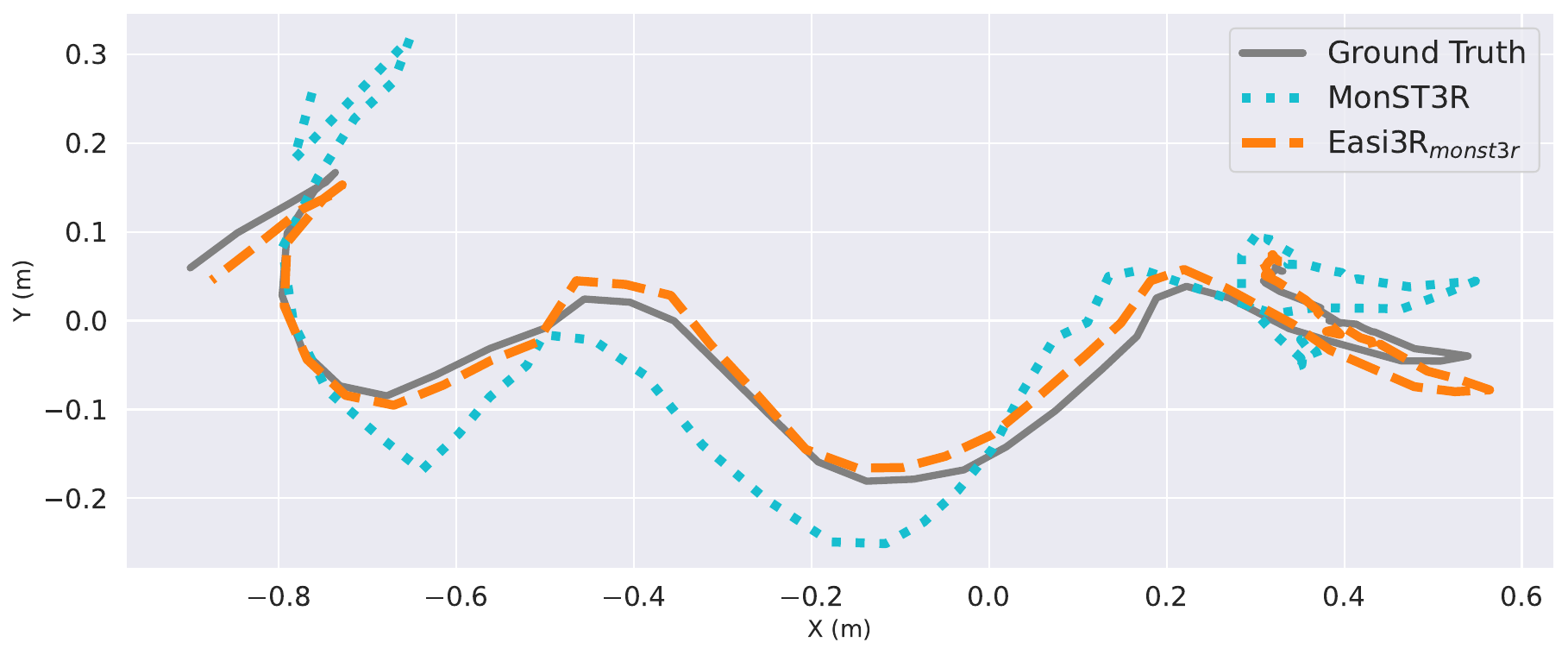}
  \end{minipage}
  \caption{{\bf Visualization of estimated camera trajectories.} 
  Our robust estimated camera trajectory (orange) deviates less from the ground truth (gray) compared to the original backbones (blue).
  }
  \label{fig:attn_traj}
\end{figure}

\begin{table}[t]
  \centering
  \scriptsize
  \renewcommand{\tabcolsep}{2pt}
  \caption{\textbf{Quantitative Comparisons of Camera Pose Estimation} on the \dycheck, \adt and \tumd datasets. The best and second best results are \textbf{bold} and \underline{underlined}, respectively. 
  }
  \label{tab:camera_pose_baseline}
  \resizebox{1.0\linewidth}{!}{
  \begin{tabular}{@{}lc|lll|lll|lll@{}}
  \toprule
  & & \multicolumn{3}{c}{\dycheck} & \multicolumn{3}{c}{\adt} & \multicolumn{3}{c}{\tumd} \\ 
  \cmidrule(lr){3-5} \cmidrule(lr){6-8} \cmidrule(lr){9-11}
  {Method} & Flow & {ATE $\downarrow$} & {RTE $\downarrow$} & {RRE $\downarrow$} & 
  {ATE $\downarrow$} & {RTE $\downarrow$} & {RRE $\downarrow$} & 
  {ATE $\downarrow$} & {RTE $\downarrow$} & {RRE $\downarrow$} \\ 
  \midrule

  \duster~\cite{dust3r} & \xmark & 0.035 & 0.030 & 2.323 & 0.042 & 0.025 & 1.212 & 0.100 & \underline{0.087} & \underline{2.692} \\
  
  \cuter~\cite{cut3r} & \xmark & \underline{0.029} & \underline{0.020} & \underline{1.383} & 0.084 & 0.025 & \textbf{0.490} & \underline{0.079} & 0.088 & 10.41 \\
  
  \monster~\cite{monst3r} & \cmark & 0.033 & 0.024 & 1.501 & 0.055 & 0.025 & 0.776 & 0.170 & 0.155 & 6.455 \\

  \daser~\cite{das3r} & \cmark & 0.033 & 0.022 & 1.467 & \underline{0.040} & 0.017 & 0.685 & 0.173 & 0.157 & 8.341 \\
  
  \textbf{\easier}$_\text{monst3r}$ & \cmark & 0.030 & 0.021 & 1.390 & \textbf{0.039} & \underline{0.016} & \underline{0.640} & 0.168 & 0.150 & 5.925 \\

  \textbf{\easier}$_\text{dust3r}$ & \cmark & \textbf{0.021} & \textbf{0.014} & \textbf{1.092} & 0.042 & \textbf{0.015} & 0.655 & \textbf{0.070} & \textbf{0.061} & \textbf{2.361} \\
  
  \bottomrule
  \end{tabular}
  }
  \end{table}

\boldparagraph{Benefits from \easier.}
We use \duster and \monster without optical flow as the backbone settings, independently analyzing the benefits that \easier offers for each. 
We show qualitative comparisons of the estimation of the camera trajectory (\figref{fig:attn_traj}) and quantitative pose accuracy in \tabref{tab:camera_pose_benefit}, including w/ and w/o flow settings.
\easier demonstrates more accurate and robust camera pose and trajectory estimation over both backbones and settings.
Our \easier effectively leverages the inherent knowledge of \duster with just a few lines of code, even achieving an improvement compared to models with optical flow prior.

\boldparagraph{Comparison.}
In \tabref{tab:camera_pose_baseline}, we compare \easier with state-of-the-art variants of \duster. Unlike the plug-and-play comparison in \tabref{tab:camera_pose_benefit}, where we optionally disable the optical flow prior for a fair evaluation. Here, we report baseline performance using their original experimental settings, \ie, whether the flow model is used is specified in the second column.
For clarity, we denote the setting ``\monster+\easier'' as \easier$_\text{monst3r}$ and ``\duster+\easier'' as \easier$_\text{dust3r}$.  
Notably, our approach achieves significant improvements and delivers the best overall performance among all methods, without ANY fine-tuning on additional dynamic datasets or mask labels.

\begin{table}[t]
  \centering
  \scriptsize
  \caption{\textbf{Benefits of Easi3R on Point Cloud Reconstruction} on the DyCheck dataset. The best and second best results are \textbf{bold} and \underline{underlined}, respectively. \easier$_\text{dust3r/monst3r}$ denotes the \easier experiment with the backbones of \monster/\duster.}
  \label{tab:point_benefit}
  \resizebox{\linewidth}{!}{
  \begin{tabular}{@{}lc|cc|cc|cc@{}}
  \toprule
  & & \multicolumn{2}{c|}{Accuracy$\downarrow$} & \multicolumn{2}{c|}{Completeness$\downarrow$} & \multicolumn{2}{c}{Distance$\downarrow$} \\
  \cmidrule(lr){3-4} \cmidrule(lr){5-6} \cmidrule(lr){7-8}
  {Method} & Flow & Mean & Median & Mean & Median & Mean & Median \\ 
  \midrule

  \duster~\cite{dust3r} & \xmark & 0.802 & \underline{0.595} & 1.950 & 0.815 & 0.353 & 0.233 \\
  \textbf{\easier}$_\text{dust3r}$ & \xmark & 0.772 & 0.596 & 1.813 & 0.757 & 0.336 & 0.219 \\

  \duster~\cite{dust3r} & \cmark & \underline{0.738} & 0.599 & \underline{1.669} & \underline{0.678} & \underline{0.313} & \underline{0.196} \\
  \textbf{\easier}$_\text{dust3r}$ & \cmark & \textbf{0.703} & \textbf{0.589} & \textbf{1.474} & \textbf{0.586} & \textbf{0.301} & \textbf{0.186} \\

  \midrule

  \monster~\cite{monst3r} & \xmark & 0.855 & 0.693 & 1.916 & 1.035 & 0.398 & 0.295 \\
  \textbf{\easier}$_\text{monst3r}$ & \xmark & \underline{0.846} & \underline{0.660} & 1.840 & 0.983 & 0.390 & 0.290 \\

  \monster~\cite{monst3r} & \cmark & 0.851 & 0.689 & \underline{1.734} & \underline{0.958} & \underline{0.353} & \textbf{0.254} \\
  \textbf{\easier}$_\text{monst3r}$ & \cmark & \textbf{0.834} & \textbf{0.643} & \textbf{1.661} & \textbf{0.916} & \textbf{0.350} & \underline{0.255} \\
  
  \bottomrule
  \end{tabular}
  }
\end{table}

\begin{table}[t]
  \centering
  \scriptsize
  \caption{\textbf{Quantitative Comparisons of Point Cloud Reconstruction} on the \dycheck dataset. The best and second best results are \textbf{bold} and \underline{underlined}, respectively. 
  }
  \label{tab:point_baseline}
  \resizebox{\linewidth}{!}{
  \begin{tabular}{@{}lc|cc|cc|cc@{}}
  \toprule
  & & \multicolumn{2}{c|}{Accuracy$\downarrow$} & \multicolumn{2}{c|}{Completeness$\downarrow$} & \multicolumn{2}{c}{Distance$\downarrow$} \\
  \cmidrule(lr){3-4} \cmidrule(lr){5-6} \cmidrule(lr){7-8}
  {Method} & Flow & Mean & Median & Mean & Median & Mean & Median \\ 
  \midrule
  
  \duster~\cite{dust3r} & \xmark & 0.802 & 0.595 & 1.950 & 0.815 & 0.353 & 0.233 \\

  \cuter~\cite{cut3r} & \xmark & \textbf{0.458} & \textbf{0.342} & \underline{1.633} & \underline{0.792} & \underline{0.326} & \underline{0.229} \\

  \monster~\cite{monst3r} & \cmark & 0.851 & 0.689 & 1.734 & 0.958 & 0.353 & 0.254 \\  

  \daser~\cite{das3r} & \cmark & 1.772 & 1.438 & 2.503 & 1.548 & 0.475 & 0.352 \\

  \textbf{\easier}$_\text{monst3r}$ & \cmark & 0.834 & 0.643 & 1.661 & 0.916 & 0.350 & 0.255 \\

  \textbf{\easier}$_\text{dust3r}$ & \cmark & \underline{0.703} & \underline{0.589} & \textbf{1.474} & \textbf{0.586} & \textbf{0.301} & \textbf{0.186} \\  
  
  \bottomrule
  \end{tabular}
  }
\end{table}

\subsection{4D Reconstruction}
\label{ssec:exp_4drecon}

We evaluate 4D reconstruction on \dycheck~\cite{dycheck} by measuring distances to ground-truth point clouds. Following prior work~\cite{Feat2GS,spann3r,aanaes2016large}, we use accuracy, completeness, and distance metrics. Accuracy is the nearest Euclidean distance from a reconstructed point to ground truth, completeness is the reverse, and distance is the Euclidean distance based on ground-truth point matching.

\boldparagraph{Quantitative Results.} 
We observe benefits from \easier in \tabref{tab:point_benefit} and \tabref{tab:point_baseline}, 
\easier demonstrates more accurate reconstruction and outperforms most baselines, even comparable to concurrent \cuter~\cite{cut3r}, which are trained with many extensive datasets.

\boldparagraph{Qualitative Results.} 
We also compare the reconstruction quality of our method with \cuter~\cite{cut3r}, \monster~\cite{monst3r} and \daser~\cite{das3r} in \figref{fig:vis_pts}. 
All baselines struggle with misalignment and entanglement of dynamic and static reconstructions, resulting in broken geometry, distortions, and ghosting artifacts.
The key to our success lies in: (1) attention-guided segmentation for robust motion disentanglement, (2) attention re-weighting for improved pairwise reconstruction, and (3) segmentation-aware global alignment for enhanced overall quality.

\section{Conclusion}

We presented \easier, an adaptation to \duster, which introduces the spatial and temporal attention mechanism behind \duster, to achieve training-free and robust 4D reconstruction.
We found the compositional complexity in attention maps, and propose a simple yet effective decomposition strategy to isolate the textureless, under-observed, and dynamic objects components and allowing for robust dynamic object segmentation.
With the segmentation, we perform a second inference pass by applying attention re-weighting, enabling robust dynamic 4D reconstruction and camera motion recovery, and at almost no additional cost on top of \duster.
Surprisingly, our experimental results demonstrate that \easier\ outperforms state-of-the-art methods in most cases.
We hope that our findings on attention map disentanglement can inspire other tasks.

\clearpage

\bigskip

\noindent\textbf{Acknowledgments.} 
We thank the members of \textit{Inception3D} and \textit{Endless AI} Labs for their help and discussions. \textit{Xingyu Chen} and \textit{Yue Chen} are funded by the Westlake Education Foundation. \textit{Xingyu Chen} is also supported by the Natural Science Foundation of Zhejiang province, China (No. QKWL25F0301). \textit{Yuliang Xiu} received funding from the Max Planck Institute for Intelligent Systems. \textit{Anpei Chen} and \textit{Andreas Geiger} are supported by the ERC Starting Grant LEGO-3D (850533) and DFG EXC number 2064/1 - project number 390727645.

{\small
\bibliographystyle{ieee_fullname}
\bibliography{bibliography,bibliography_long,bibliography_custom}
}

\clearpage
\appendix
\setcounter{page}{1}
\maketitlesupplementary

In this \textbf{supplementary document}, we first present additional method details on temporal consistency dynamic object segmentation in~\cref{sec:temporal}.
Next, we conduct ablation studies of \easier in~\cref{sec:ablation} and analysis limitations in~\cref{sec:limitation}.
Lastly, we report additional qualitative results in~\cref{sec:addtional_results}. We invite readers to \href{https://easi3r.github.io/}{easi3r.github.io} for better visualization.

\section{Dynamic Object Segmentation}
\label{sec:temporal}
We have presented dynamic object segmentation for a single frame in~\secref{ssec:dynamic}, now we introduce how to ensure consistency along the temporal axis.
Given image feature tokens $\bF_0^t$ for frames at $t$, output from the image encoder, we concatenate them along the temporal dimension,
\begin{equation}
  \small
  \bar{\bF} = [\bF_0^1; \bF_0^2; \dots; \bF_0^T] \in \mathbb{R}^{(T \times h \times w) \times c}
\end{equation}
where $c$ is the feature dimension of the tokens. This allows us to apply k-means clustering to group similar features across frames, producing cluster assignments,
\begin{equation}
  \small
  C = \text{KMeans}(\bar{\bF}, k), \quad C^t(x,y) \in \{1,\ldots,k\},\,\, \forall t, x, y
\end{equation}
where $k$ is the number of clusters, we use $k=64$ for all experiments. 

For each cluster $c \in \{1,\ldots,k\}$, we compute a dynamic score $s_c$ by averaging the base dynamic attention values of all tokens within that cluster:
\begin{equation}
  s_c = \frac{\sum_{t}\sum_{i,j} \mathbbm{1}[C^t(x,y)=c] \cdot \bA^{t=\DYN}(x,y)}{\sum_{t}\sum_{x,y} \mathbbm{1}[C^t(x,y)=c]}
\end{equation}
where $\mathbbm{1}[\cdot]$ denotes the indicator function.
We then use these scores to generate a cluster-fused dynamic attention map, mapping each pixel’s cluster assignment back to its corresponding dynamic score,
\begin{equation}
  \bA^{t=\DYN}_{\text{fuse}}(x,y) = s_{C^t(x,y)}
\end{equation}

The refined dynamic attention map $\bA^{t=\DYN}_{\text{fuse}} \in \mathbb{R}^{h \times w}$ is used to infer the dynamic object segmentation by,
\begin{equation}
  \bM^{t}(x,y) = \mathbbm{1} [\bA^{t=\DYN}_{\text{fuse}}(x,y) > \alpha ]
\end{equation}
where $\alpha$ is an automatic image thresholding using \href{https://en.wikipedia.org/wiki/Otsu%27s_method}{Otsu's method~\cite{Otsu}}.
The resulting dynamic object segmentation is further utilized in the second inference pass and global optimization.

\begin{figure}[t!]
    \centering
    \includegraphics[width=1\linewidth,page=1]{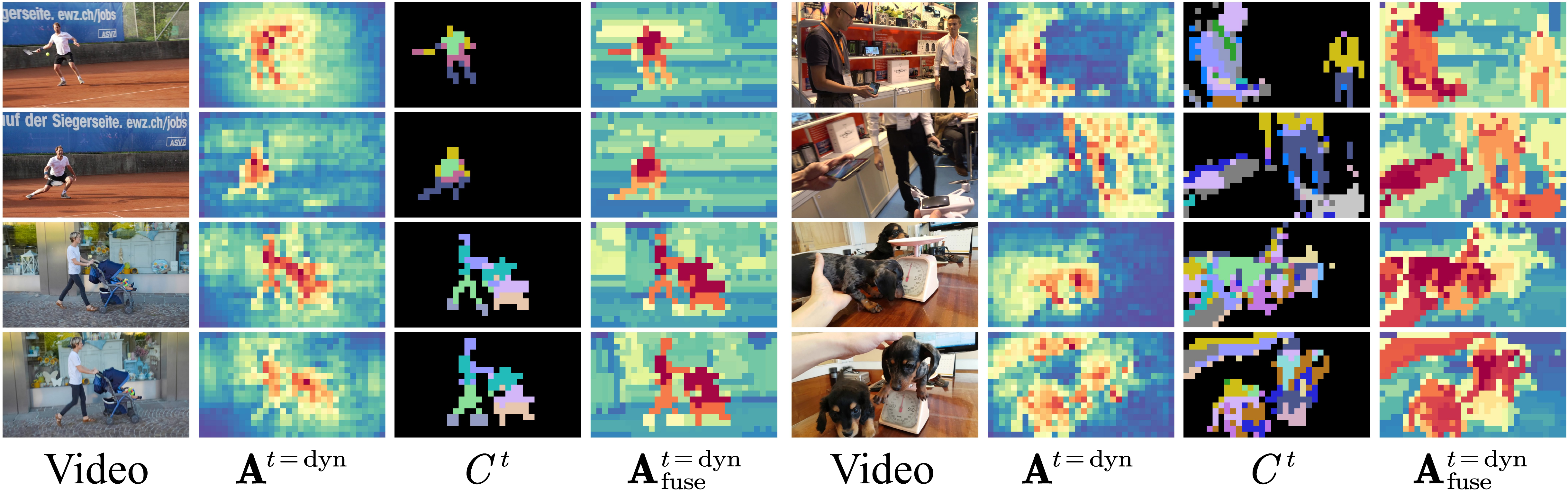}
    \caption{{\bf Benefits of Cross-frame Feature Clustering.} We visualize the dynamic attention map $\bA^{t=\DYN}$, cluster assignments $C^{t}$, and cluster-fused dynamic attention map $\bA^{t=\DYN}_{\text{fuse}}$.
    Features from the DUSt3R encoder exhibit temporal consistency, as cluster assignments ($C^{t}$) remain unchanged across frames, thereby enhancing temporal consistency in dynamic segmentation ($\bA^{t=\DYN}{\text{fuse}}$) through clustering-guided temporal fusing. For better visual intuition, we invite readers to \href{https://easi3r.github.io/}{easi3r.github.io}.
    }
    \label{fig:cluster}
  \end{figure}

\begin{table}[t]
  \caption{\textbf{Ablation of Dynamic Object Segmentation} on \davis.}
  \label{tab:exp_dyn_seg_ablation}
  \resizebox{1.0\linewidth}{!}{
  \begin{tabular}{ll|ll|ll|ll} 
  \toprule
  & & \multicolumn{2}{c|}{\davis-16} & \multicolumn{2}{c|}{\davis-17} & \multicolumn{2}{c}{\davis-all} \\ 
   
  \cmidrule(lr){3-4} \cmidrule(lr){5-6} \cmidrule(lr){7-8} 
  
  Backbone & Ablation & {\JM $\uparrow$} & {\JR $\uparrow$} & {\JM $\uparrow$} & {\JR $\uparrow$} & {\JM $\uparrow$} & {\JR $\uparrow$}\\ 
  \midrule
  
  \multirow{6}{*}{\duster}
  & w/o $\bA^{a=\SRC}_{\mu}$ & 45.1 & 45.2 & 42.8 & 39.9 & 42.2 & 38.5 \\
  
  & w/o $\bA^{a=\SRC}_{\sigma}$ & 42.3 & 50.0 & 35.0 & 37.0 & 30.9 & 28.3  \\
  
  & w/o $\bA^{a=\REF}_{\mu}$ & 33.3 & 28.4 & 31.5 & 27.9 & 32.5 & 29.7 \\

  & w/o $\bA^{a=\REF}_{\sigma}$ & 47.7 & 54.1 & 46.2 & 54.3 & 43.7 & 48.6 \\

  & w/o Clustering & 40.0 & 38.5 & 38.3 & 38.3 & 34.3 & 30.5 \\

  & Full & \textbf{53.1} & \textbf{60.4} & \textbf{49.0} & \textbf{56.4} & \textbf{44.5} & \textbf{49.6} \\

  \midrule

  \multirow{6}{*}{\monster}
  
  & w/o $\bA^{a=\SRC}_{\mu}$ & 47.2 & 51.5 & 44.4 & 46.7 & 40.9 & 41.5 \\
  
  & w/o $\bA^{a=\SRC}_{\sigma}$ & 49.7 & 60.1 & 48.7 & 57.8 & 44.9 & 49.6 \\
  
  & w/o $\bA^{a=\REF}_{\mu}$ & 46.4 & 54.0 & 47.4 & 55.9 & 45.3 & 50.7 \\

  & w/o $\bA^{a=\REF}_{\sigma}$ & 50.7 & 62.6 & 51.0 & 60.2 & 50.3 & 56.8 \\

  & w/o Clustering & 45.5 & 46.7 & 45.3 & 48.1 & 42.1 & 43.5 \\

  & Full & \textbf{57.7} & \textbf{71.6} & \textbf{56.5} & \textbf{68.6} & \textbf{53.0} & \textbf{63.4} \\
  \bottomrule
  \end{tabular}
  }
\end{table}

\section{Ablation Study}
\label{sec:ablation}
Our ablation lies in two folds: dynamic object segmentation and 4D reconstruction.
For dynamic object segmentation, as shown in \tabref{tab:exp_dyn_seg_ablation} we ablate the contribution of four aggregated temporal cross-attention maps, $\bA^{a=\SRC}_{\mu},\bA^{a=\SRC}_{\sigma},\bA^{a=\REF}_{\mu},\bA^{a=\REF}_{\sigma}$, and feature clustering.
The ablation results show that (1) Disabling any temporal cross-attention map leads to a performance drop, indicating that all attention maps contribute to improved dynamic object segmentation;
and (2) Features from the \duster encoder exhibit temporal consistency and enhance dynamic segmentation through cross-frame clustering.

\begin{table*}[t]
  \centering
  \scriptsize
  \caption{\textbf{Ablation Study of Camera Pose Estimation and Point Cloud Reconstruction} on the \dycheck dataset.}
  \label{tab:rec_ablation}
  \resizebox{\textwidth}{!}{
  \begin{tabular}{@{}ccc|ccc|cc|cc|cc@{}}
  \toprule
  & & & \multicolumn{3}{c|}{Pose Estimation} & \multicolumn{6}{c}{Reconstruction} \\
  \cmidrule(lr){4-6} \cmidrule(lr){7-12}
  & & & \multirow{2}{*}{ATE$\downarrow$} & \multirow{2}{*}{RTE$\downarrow$} & \multirow{2}{*}{RRE$\downarrow$} & \multicolumn{2}{c|}{Accuracy$\downarrow$} & \multicolumn{2}{c|}{Completeness$\downarrow$} & \multicolumn{2}{c}{Distance$\downarrow$} \\ 
  Backbone & {Re-weighting} & Flow-GA & & & & Mean & Median & Mean & Median & Mean & Median \\ 
  \midrule

  \multirow{4}{*}{\duster}

  & Ref + Src & \xmark & 0.030 & 0.026 & 1.777 & 0.775 & 0.596 & 1.848 & 0.778 & 0.342 & 0.224 \\
        & Ref & \xmark & 0.029 & 0.025 & 1.774 & 0.772 & 0.596 & 1.813 & 0.757 & 0.336 & 0.219 \\

  \cmidrule{2-12}

  & Ref & w/o Mask & 0.026 & 0.017 & 1.472 & 0.940 & 0.831 & 1.654 & 0.685 & 0.336 & 0.220 \\
  & Ref & w/ Mask  & \textbf{0.021} & \textbf{0.014} & \textbf{1.092} & \textbf{0.703} & \textbf{0.589} & \textbf{1.474} & \textbf{0.586} & \textbf{0.301} & \textbf{0.186} \\

  \addlinespace[1.5pt]
  \hline\hline
  \addlinespace[1.5pt]

  \multirow{4}{*}{\monster}
  & Ref + Src & \xmark & 0.040 & 0.032 & 1.751 & 0.848 & 0.744 & 1.850 & 1.003 & 0.398 & 0.292 \\
        & Ref & \xmark & 0.038 & 0.032 & 1.736 & 0.846 & 0.660 & 1.840 & 0.983 & 0.390 & 0.290 \\

  \cmidrule{2-12}

  & Ref & w/o Mask & 0.033 & 0.023 & 1.495 & 0.969 & 0.796 & 1.752 & 0.998 & 0.368 & 0.273 \\
  & Ref & w/ Mask  & \textbf{0.030} & \textbf{0.021} & \textbf{1.390} & \textbf{0.834} & \textbf{0.643} & \textbf{1.661} & \textbf{0.916} & \textbf{0.350} & \textbf{0.255} \\
  
  \bottomrule
  \end{tabular}
  }
\end{table*}

\tabref{tab:rec_ablation} presents ablation studies on 4D reconstruction, evaluating two key design choices: (1) the impact of two-branch re-weighting (applying attention re-weighting to both reference and source decoders) and (2) global alignment using optical flow with and without segmentation.
The ablation results show that 
(1) Re-weighting only the reference view decoder outperforms re-weighting both branches. Since the reference and source decoders serve different roles, and the reference view acts as the static standard, this aligns with our design intuition (\rmnum{1});
and (2) Incorporating segmentation in global alignment consistently improves 4D reconstruction quality.

\section{Limitations}
\label{sec:limitation}
\begin{figure}[t!]
    \centering
    \includegraphics[width=1\linewidth,page=1]{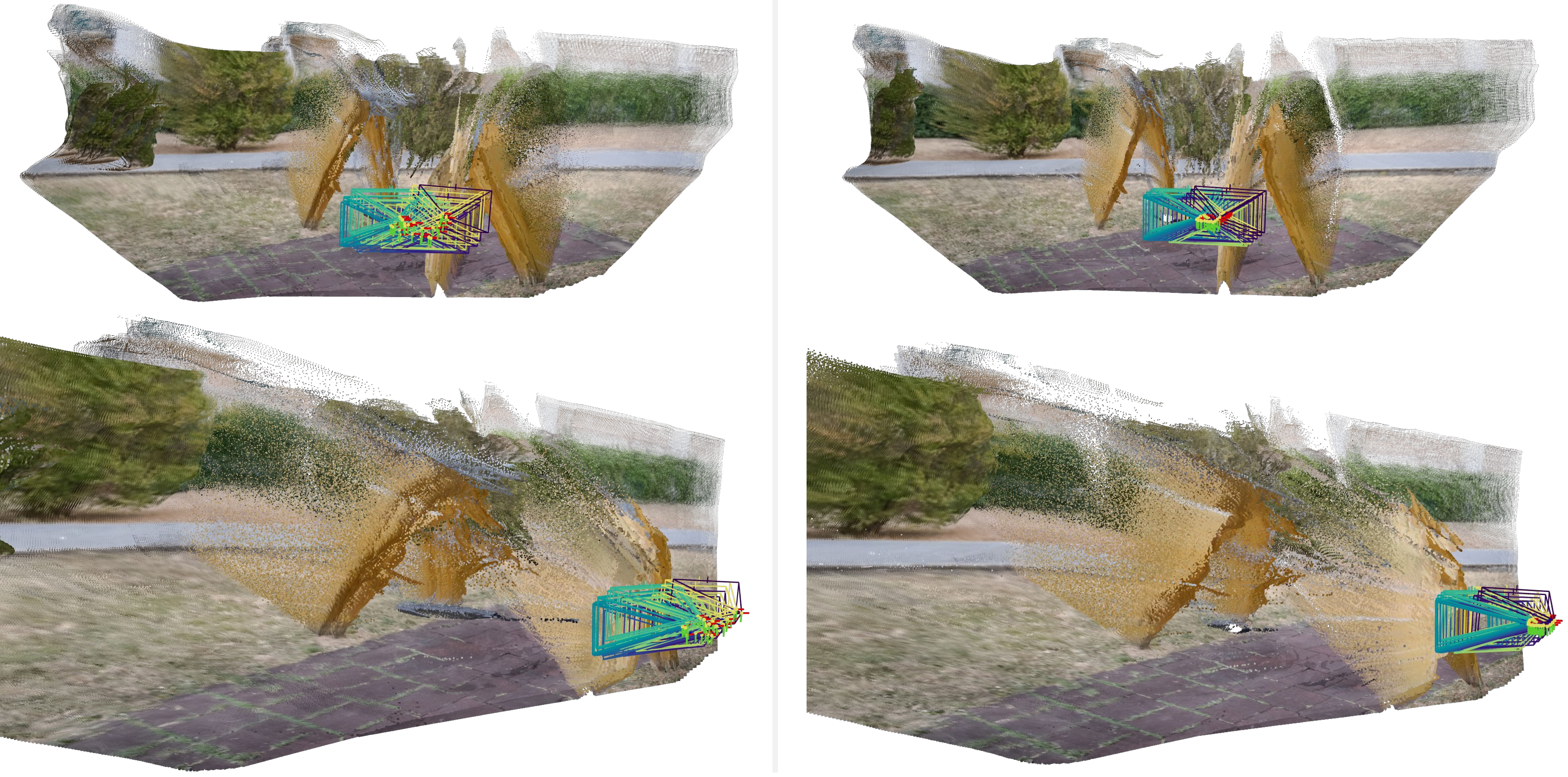}
     \setlength{\tabcolsep}{0.05pt}     %
       \small
          \begin{tabularx}{\linewidth}{
             >{\centering\arraybackslash}m{0.5\linewidth}
             >{\centering\arraybackslash}m{0.5\linewidth}
            }
            \duster~\cite{dust3r} & Ours
             \end{tabularx}
    \caption{{\bf Limitation.} We visualize static reconstructions from two different viewpoints in the top and bottom rows.  
    \easier improves camera pose estimation and point cloud reconstruction (top row), enhancing alignment in structures like swing supports through attention re-weighting and segmentation-aware global alignment. However, from another viewpoint (bottom row), \easier still produces floaters near object boundaries.}
    \label{fig:limitation}
  \end{figure}

Despite strong performance on various in-the-wild videos, \easier can fail when the \duster/\monster backbones produce inaccurate depth predictions.
While \easier effectively improves camera pose estimation and point cloud reconstruction, as shown in \tabref{tab:point_baseline} of the main paper, it provides clear improvements in completeness and distance metrics, which are measured on the global point cloud. However, a noticeable gap remains in depth accuracy, which is evaluated on per-view outputs. This is because our method focuses mainly on improving dynamic regions and global alignment rather than correcting depth predictions in static parts, as illustrated in~\figref{fig:limitation}.
We leave per-view depth correction for future work.

\begin{figure}[t!]
    \centering
    \includegraphics[width=1\linewidth,page=1]{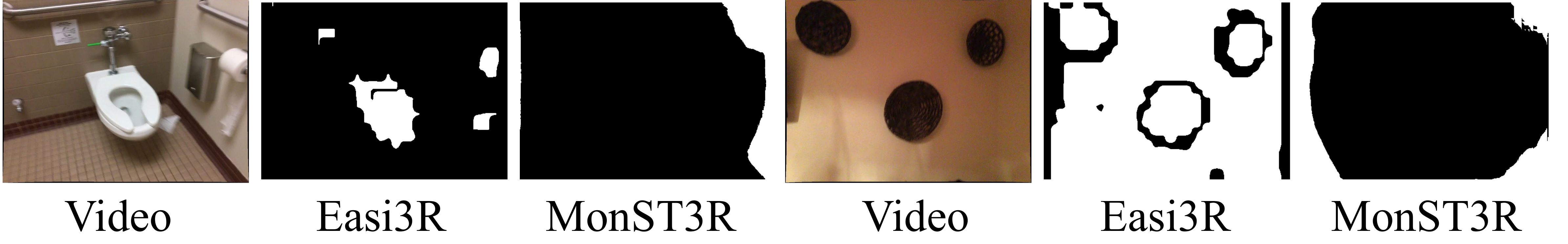}
    \vspace{-20pt}
    \caption{{\bf Dynamic masks in static scene.} 
    \easier tends to reweight low confident regions in static scenes, leading to better pose estimation, as shown in \cref{tab:video_pose_supp}.
    }
    \label{fig:static}
  \end{figure}
\begin{table}[t]
\vspace{-6pt}
\centering
\renewcommand{\arraystretch}{1.02}
\renewcommand{\tabcolsep}{1.5pt}
\resizebox{\linewidth}{!}{
\begin{tabular}{@{}ll>{\centering\arraybackslash}p{1.5cm}>{\centering\arraybackslash}p{1.5cm}|>{\centering\arraybackslash}p{1.5cm}>{\centering\arraybackslash}p{1.5cm}|>{\centering\arraybackslash}p{1.5cm}>{\centering\arraybackslash}p{1.5cm}@{}}
\toprule
 &  & \multicolumn{2}{c}{\textbf{Sintel}} & \multicolumn{2}{c}{\textbf{BONN}} & \multicolumn{2}{c}{\textbf{KITTI}} \\ 
\cmidrule(lr){3-4} \cmidrule(lr){5-6} \cmidrule(lr){7-8}
\textbf{Alignment} & \textbf{Method} & {Abs Rel $\downarrow$} & {$\delta$\textless $1.25\uparrow$} & {Abs Rel $\downarrow$} & {$\delta$\textless $1.25\uparrow$} & {Abs Rel $\downarrow$} & {$\delta$ \textless $1.25\uparrow$} \\ 
\midrule
\multirow{15}{*}{\begin{minipage}{3cm}Per-sequence \\
scale  \& shift\end{minipage}} 
 & Marigold & 0.532 & {51.5} & {0.091} & {93.1} & {0.149} & {79.6} \\
  & DepthAnythingV2 & 0.367 & {55.4} & {0.106} & {92.1} &{0.140} & {80.4} \\
   & NVDS & 0.408 & {48.3} & {0.167} & {76.6} & {0.253} & {58.8} \\
    & ChronoDepth & 0.687 & {48.6} & {0.100} & {91.1} & {0.167} &{75.9} \\
     & DepthCrafter & \textbf{0.292} & \textbf{{69.7}} & {0.075} & \underline{{97.1}} & 0.110 & {88.1} \\
      & Robust-CVD & 0.703 & {47.8} & {-} & {-} & - & - \\
      & CasualSAM & 0.387 & {54.7} & {0.169} & {73.7} & {0.246} & 62.2 \\

& MASt3R & 0.327 & \underline{59.4} & {0.167} & {78.5} & {0.137} & {83.6} \\

& Spann3R & 0.508 & {50.8} & {0.157} & {82.1} & {0.207} & {73.0} \\

& CUT3R & 0.454  & 55.7 & 0.074 & {94.5} & \underline{0.106} & \underline{88.7} \\

& DUSt3R & 0.531 & {51.2} & {0.156} & {83.1} & {0.135} & {81.8} \\
& MonST3R & {0.333} & {59.0} & \underline{{0.066}} & 96.4 & {0.157} & {73.8} \\

& \textbf{\easier}$_\text{dust3r}$ & 0.435  & 59.1 & 0.085 & 91.1 & 0.155 & 76.1 \\
& \textbf{\easier}$_\text{monst3r}$ & \underline{0.316}  & 59.3 & \textbf{0.057} & \textbf{97.2} & \textbf{0.092} & \textbf{90.6} \\

\midrule
\multirow{8}{*}{\begin{minipage}{3cm}Per-sequence scale\end{minipage}}
& MASt3R & 0.641 & {43.9} & {0.252} & {70.1} & {0.183} & {74.5} \\

& Spann3R & 0.622 & {42.6} & {0.144} & {81.3} & {0.198} & {73.7} \\

& CUT3R & 0.421  & 47.9 & 0.078 & 93.7 & \underline{0.118} & \underline{88.1} \\

& DUSt3R & 0.656 & {45.2} & {0.155} & {83.3} & 0.144 & 81.3 \\
& MonST3R & \underline{0.378} & \underline{55.8} & \underline{0.067} & \underline{96.3} & {0.168} & {74.4} \\

& \textbf{\easier}$_\text{dust3r}$ & 0.577  & 51.9 & 0.086 & 90.3 & 0.170 & 74.2 \\
& \textbf{\easier}$_\text{monst3r}$ & \textbf{0.377}  & \textbf{55.9} & \textbf{0.059} & \textbf{97.0} & \textbf{0.102} & \textbf{91.2} \\

\bottomrule
\end{tabular}
}
\vspace{-4pt}
\caption{\small{
\textbf{Video Depth Evaluation}. 
We use the evaluation results from CUT3R for baselines.
}
\vspace{-8pt}
}
\label{tab:video_depth_supp}
\end{table}

\begin{table}[t]
\centering
\footnotesize
\renewcommand{\arraystretch}{1.}
\renewcommand{\tabcolsep}{2.5pt}
\resizebox{\linewidth}{!}{
\begin{tabular}{@{}llccc|ccc|ccc@{}}
\toprule
& & \multicolumn{3}{c}{\textbf{Sintel}} & \multicolumn{3}{c}{\textbf{TUM-dynamics}} & \multicolumn{3}{c}{\textbf{ScanNet (static)}} \\ 
\cmidrule(lr){3-5} \cmidrule(lr){6-8} \cmidrule(lr){9-11}
{\textbf{Category}} & {\textbf{Method}} & {ATE $\downarrow$} & {RPE trans $\downarrow$} & {RPE rot $\downarrow$} & {ATE $\downarrow$} & {RPE trans $\downarrow$} & {RPE rot $\downarrow$} & {ATE $\downarrow$} & {RPE trans $\downarrow$} & {RPE rot $\downarrow$} \\ 
\midrule
\multirow{4}{*}{Pose only} & DROID-SLAM  & 0.175 & 0.084 & {1.912} & - & - & - & - & - & - \\ 
& DPVO  & \underline{0.115} & {0.072} & 1.975 & - & - & - & - & - & - \\ 
& Particle-SfM & 0.129 & \textbf{0.031} & \bf{0.535} & - & - & - & 0.136 & 0.023 & 0.836 \\ 
& LEAP-VO  & \textbf{{0.089}} & \underline{0.066} & \underline{1.250} & {{0.068}} & {0.008} & {1.686} & {\textbf{0.070}} & {\textbf{0.018}} & {\textbf{0.535}} \\ 

\midrule

\multirow{9}{*}{Pose  \& Depth} & Robust-CVD & 0.360 & 0.154 & 3.443 & 0.153 & 0.026 & 3.528 & 0.227 & 0.064 & 7.374 \\ 
 & CasualSAM & 0.141 & \textbf{0.035} & \textbf{0.615} & {0.071} & \textbf{0.010} & 1.712 & 0.158 & 0.034 & 1.618 \\ 
 
{ }& MASt3R & {{0.185}} & {0.060} & {1.496} & {\bf{0.038}} & {\underline{0.012}} & {\bf{0.448}} & {{0.078}} & {{0.020}} & {\bf {0.475}} \\

& Spann3R & {{0.329}} & {0.110} & {4.471} & {{0.056}} & {{0.021}} & {{0.591}} & {{0.096}} & {{0.023}} & {{0.661}} \\ 

& CUT3R &  0.213 & 0.066 & \underline{0.621} & \underline{0.046} & 0.015 & \underline{0.473} & {0.099} & {0.022} & {0.600}\\  

& DUSt3R & 0.417 & 0.250 & 5.796 & 0.083 & 0.017 & 3.567 & {0.081} & 0.028 & 0.784 \\ 
& MonST3R  & \underline{{0.111}} & {0.044} & {0.869} & {{0.098}} & {{0.019}} & {{0.935}} & {{0.077}} & {\underline{0.018}} & {{0.529}} \\

& \textbf{\easier}$_\text{dust3r}$ & 0.402 & 0.098 & 0.876 & 0.134 & 0.017 & 1.077 & \underline{0.067} & \underline{0.018} & 0.670 \\ 
& \textbf{\easier}$_\text{monst3r}$ & \textbf{0.110}  & \underline{0.042} & 0.758 & 0.105 & 0.022 & 1.064 & \textbf{0.061} & \textbf{0.017} & \underline{0.525} \\

\bottomrule
\end{tabular}
}
\vspace{-4pt}
\caption{\textbf{Camera Pose Evaluation.} 
We use the evaluation results from CUT3R for baselines.
}
\label{tab:video_pose_supp}
\end{table}

\begin{table}[t]
  \resizebox{\linewidth}{!}{
  \begin{tabular}{llccc|cc|cc|cc} 
  \toprule
  & & \multicolumn{3}{c}{\textbf{Input}} & \multicolumn{2}{c}{\textbf{Output}} & \multicolumn{2}{c}{\textbf{\davis-16}} & \multicolumn{2}{c}{\textbf{\davis-17}} \\ 
   
  \cmidrule(lr){3-5} \cmidrule(lr){6-7} \cmidrule(lr){8-9} \cmidrule(lr){10-11}
  
  \textbf{Supervision} & \textbf{Method} & RGB & Optical Flow & Point Tracks & Mask & 4D Reconstruction & {\JM $\uparrow$} & {\JM-M $\uparrow$} & {\JM $\uparrow$} & {\JM-M $\uparrow$} \\ 
  \midrule
  
  \multirow{7}{*}{Supervised}
  & SFL & \checkmark & \checkmark &  & \checkmark &  & 67.4 & - &  - & -\\
  & SIMO  & \checkmark & \checkmark &  & \checkmark &  & 67.8 & - &  - & -\\

  & OCLR-flow  & \checkmark & \checkmark &  & \checkmark &  & 72.0  & 70.0 & -  & 69.9\\
  & OCLR-TTA  & \checkmark & \checkmark &  & \checkmark &  &  80.8 & 80.2 & -  & \underline{76.0}\\
  & FlowSAM  & \checkmark & \checkmark &  & \checkmark &  & \underline{87.1} & \underline{85.7} &  - & -\\
  & SegAnyMo  & \checkmark &  & \checkmark & \checkmark &  & \textbf{90.6} & \textbf{89.2} & -  & \textbf{90.0}\\
  & DAS3R  & \checkmark &  &  & \checkmark & \checkmark & 54.2 & 51.6 & 57.4 & 55.5 \\ 
  
  \midrule

  \multirow{9}{*}{Unsupervised}
  & SAGE & \checkmark & \checkmark &  & \checkmark &  & 42.6 & - &  - & -\\
  & CUT  & \checkmark & \checkmark &  & \checkmark &  & 55.2 & - &  - & -\\
  & FTS  & \checkmark & \checkmark &  & \checkmark &  &  55.8 & - &  - & -\\
  & CIS  & \checkmark & \checkmark &  & \checkmark &  & 70.3  & 67.6 & - & -\\
  & Motion Grouping & \checkmark & \checkmark &  & \checkmark &  & 68.3 & - &  - & -\\
  & EM  & \checkmark & \checkmark & & \checkmark &  & 69.3 & 76.2 & -  & -\\
  & RCF-Stage1  & \checkmark & \checkmark & & \checkmark &  &  80.2 & \underline{78.6} & -  & -\\
  & RCF-All   & \checkmark & \checkmark & & \checkmark &  & \underline{82.1} & \textbf{81.0} & - & -\\
  & LRTL   & \checkmark & \checkmark & \checkmark & \checkmark &  & \textbf{82.2} & - & - & -\\
  
  \midrule

  \multirow{3}{*}{Zero-shot}
  
  & MonST3R & \checkmark & \checkmark &  & \checkmark & \checkmark & 64.3 & 61.4 & 56.4 & 59.0 \\

  & \textbf{\easier}$_\text{dust3r}$ & \checkmark &  &  & \checkmark & \checkmark &  \underline{67.9} & \underline{67.4} & \underline{60.1} & \underline{62.0}\\
  & \textbf{\easier}$_\text{monst3r}$ & \checkmark &  &  & \checkmark & \checkmark & \textbf{70.7} & \textbf{71.1} & \textbf{67.9} & \textbf{67.7}\\
  \bottomrule
  \end{tabular}
  }
  \vspace{-4pt}
  \caption{\textbf{Comparisons of Dynamic Object Segmentation} on \davis with 2D dynamic segmentation methods.}
\label{tab:exp_dyn_seg_2d}
\end{table}
\begin{table}[t]
\centering
  \resizebox{0.98\linewidth}{!}{
  \begin{tabular}{lcll|ll|ll} 
  \toprule
  & & \multicolumn{2}{c}{\textbf{\davis-16}} & \multicolumn{2}{c}{\textbf{\davis-17}} & \multicolumn{2}{c}{\textbf{\davis-all}} \\ 
   
  \cmidrule(lr){3-4} \cmidrule(lr){5-6} \cmidrule(lr){7-8} 
  
  \textbf{Ablation} & \textbf{Variants} & {\JM $\uparrow$} & {\JR $\uparrow$} & {\JM $\uparrow$} & {\JR $\uparrow$} & {\JM $\uparrow$} & {\JR $\uparrow$}\\ 
  \midrule
  
  \multirow{3}{*}{Window Size}
  
  & 3 & 76.0 & 89.2 & 70.8 & 82.4 & 65.7 & 76.2 \\
  
  & 5* & 70.7 & 79.9 & 67.9 & 76.1 & 63.1 & 72.6 \\
  
  & 7 & 66.9 & 76.9 & 63.9 & 73.3 & 61.0 & 68.8 \\

  \midrule

  \multirow{4}{*}{Number of Clusters}

  & 16 & 67.4 & 79.1 & 64.6 & 73.7 & 60.7 & 65.2 \\
  
  & 32 & 71.6 & 83.9 & 68.0 & 78.3 & 65.1 & 75.2 \\
  
  & 64* & 70.7 & 79.9 & 67.9 & 76.1 & 63.1 & 72.6 \\
  
  & 128 & 69.9 & 79.9 & 66.3 & 76.2 & 62.9 & 73.1 \\
  
  \midrule

  \multirow{3}{*}{Thresholding Values}
  
  % & 0.3 & 38.9 & 30.7 & 42.5 & 38.4 & 41.8 & 38.6 \\
  & 0.5 & 61.6 & 64.5 & 61.0 & 65.2 & 61.6 & 67.8 \\
  & 0.7 & 70.2 & 85.0 & 62.8 & 71.8 & 58.1 & 67.0 \\
  & Otsu's method* & 70.7 & 79.9 & 67.9 & 76.1 & 63.1 & 72.6 \\
  
  \bottomrule
  \end{tabular}
  }
  \vspace{-4pt}
  \caption{\textbf{More ablations on segmentation quality using \davis.} * denotes the value used in the submission.}
  \vspace{-8pt}
  \label{tab:exp_dyn_seg_ablation}
\end{table}

\section{Addtional Results}
\boldparagraph{More Evaluation.}
We evaluated on full-length sequences with downsampling only for GPU fit, leading to more dynamic and challenging motion than prior short-clip settings. For a more general evaluation, we also include \monster and \cuter evaluation protocols. \cref{tab:video_depth_supp} and \cref{tab:video_pose_supp} confirm the effectiveness. 

\boldparagraph{Behavior for Static Scenes.}
Interestingly, our method also improves on static scenes (ScanNet in \cref{tab:video_pose_supp} and \cref{fig:static}), owing to our attention reweighting.

\boldparagraph{Comparison with 2D Baselines.} 
We also include a comparison with 2D baselines. As shown in \cref{tab:exp_dyn_seg_2d}, Easi3R achieves SOTA segmentation in a zero-shot manner with only the image as input.

\boldparagraph{More Ablation.}
We further ablate the default settings - window size of 5, 64 clusters, and Otsu's method in \cref{tab:exp_dyn_seg_ablation}. Using the recent segmentation method SegAnyMo, pose accuracy improves by 9.62\% and depth accuracy improves by 4.11\% on the Sintel dataset.

\boldparagraph{Runtime}
Our method runs at almost the same speed as \monster. \monster runs at $0.33$ FPS, while Easi3R achieves $0.31$ FPS for $512 \times 144$ image resolution on an NVIDIA RTX 4090 GPU.

\begin{figure*}[t!]
    \centering
      \small
     \renewcommand{\arraystretch}{0.5} %
     \setlength{\tabcolsep}{0.005pt}     %
  \begin{tabularx}{\linewidth}{
     >{\centering\arraybackslash}m{0.05\linewidth}
     >{\centering\arraybackslash}m{0.2375\linewidth}
     >{\centering\arraybackslash}m{0.2375\linewidth}
     >{\centering\arraybackslash}m{0.2375\linewidth}
     >{\centering\arraybackslash}m{0.2375\linewidth}
    }
    Video & \quad \cuter~\cite{cut3r} & \monster~\cite{monst3r} & \daser~\cite{das3r} & Ours 
     \end{tabularx}
    \includegraphics[width=1\linewidth,page=1]{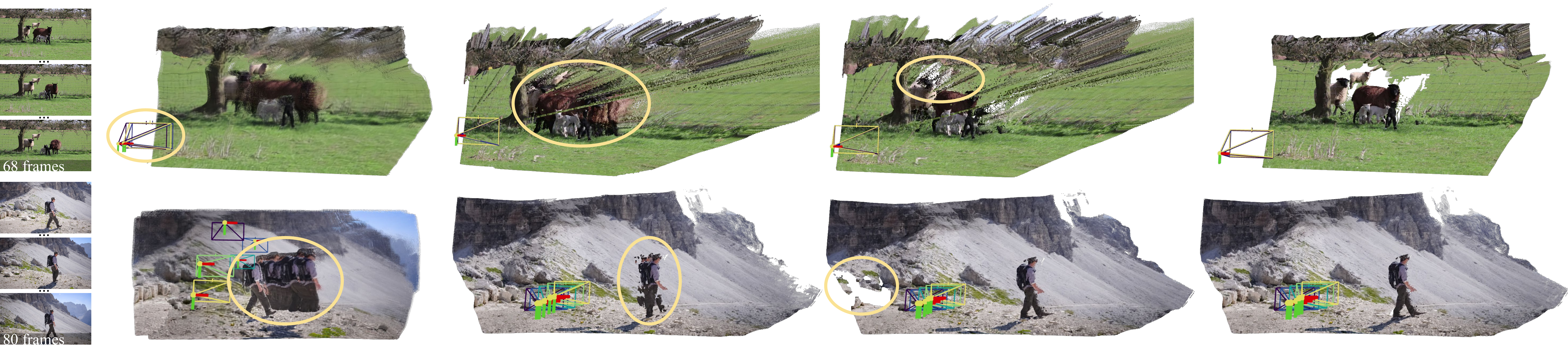}
    \caption{{\bf Qualitative Comparison.} 
    We visualize cross-frame globally aligned static scenes with dynamic point clouds at a selected timestamp. Notably, instead of using ground truth dynamic masks in previous work, we apply the estimated per-frame dynamic masks to filter out dynamic points at other timestamps for comparison. Top and bottom rows are \easier$_\text{dust3r/monst3r}$, respectively.}
    \vspace{-6pt}
    \label{fig:add_sota}
  \end{figure*}

\begin{figure*}[t!]
    \centering
    \includegraphics[width=1\linewidth,page=1]{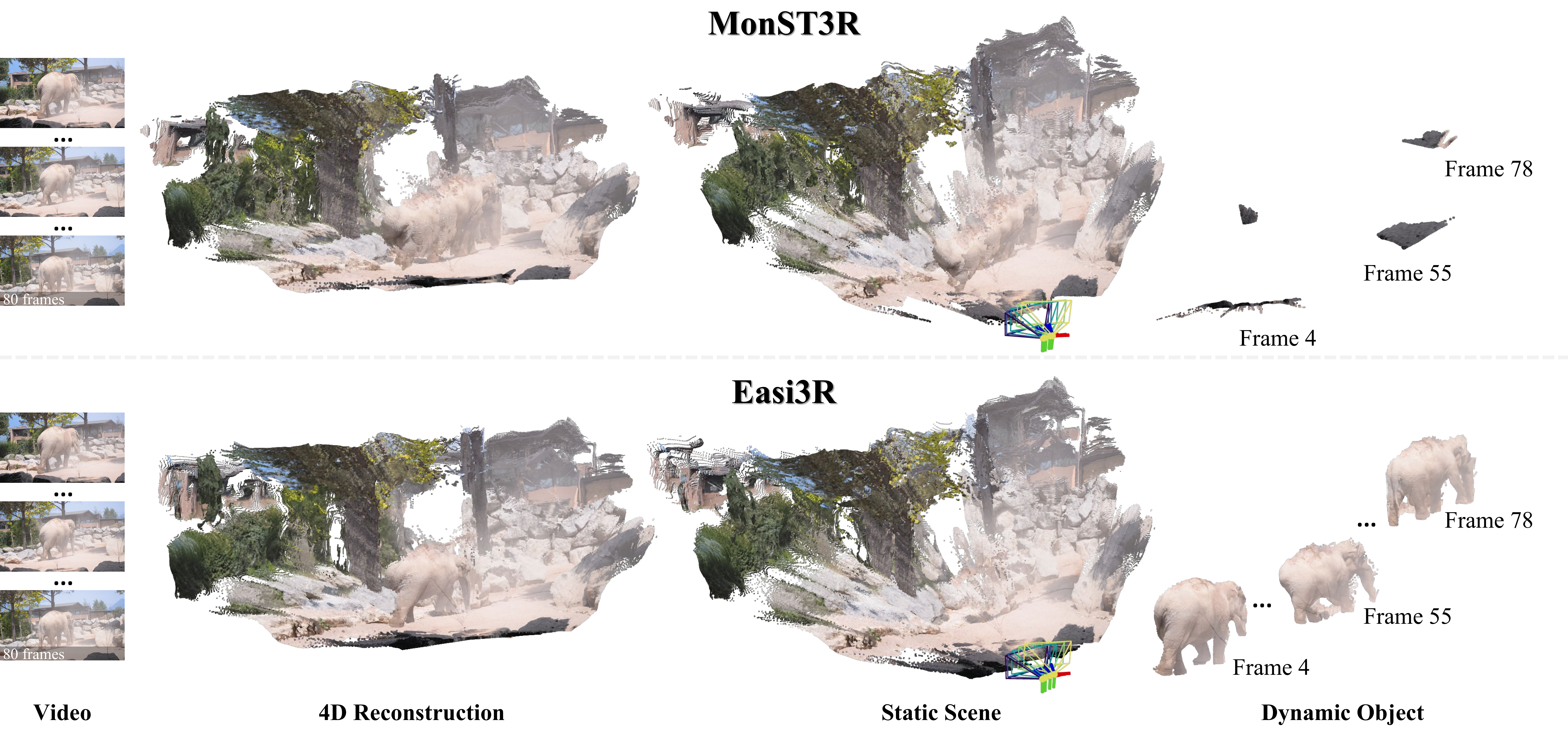}
    \caption{{\bf Disentanglement \vs \monster~\cite{monst3r}.}
    We visualize the disentangled 4D reconstruction, static scene and dynamic objects at different frames. \monster tends to predict under-segmented dynamic masks.
    }
    \vspace{-6pt}
    \label{fig:vs_monst3r}
  \end{figure*}

\begin{figure*}[t!]
    \centering
    \includegraphics[width=1\linewidth,page=1]{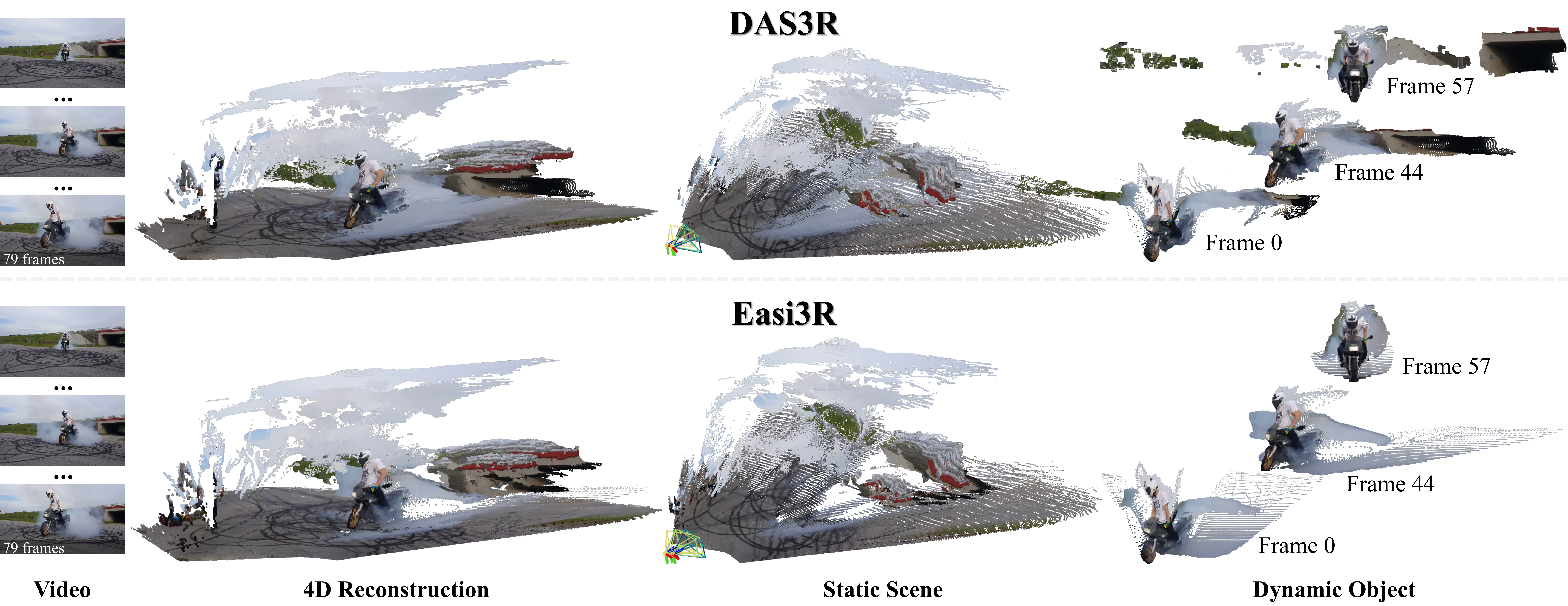}
    \caption{{\bf Disentanglement \vs \daser~\cite{das3r}.}
    We visualize the disentangled 4D reconstruction, static scene and dynamic objects at different frames. \daser tends to predict over-segmented dynamic masks.}
    \vspace{-18pt}
    \label{fig:vs_das3r}
  \end{figure*}

\boldparagraph{Qualitative Results}
We report additional qualitative results of disentangled 4D reconstruction in \figref{fig:add_sota}, \figref{fig:vs_monst3r} and \figref{fig:vs_das3r}. 
We find that \monster tends to predict under-segmented dynamic masks, while \daser tends to predict over-segmented dynamic masks. \cuter, although it produces more accurate depth estimation, is prone to being affected by dynamic objects, leading to misaligned static structures, unstable camera pose estimation, and ghosting artifacts due to the lack of dynamic segmentation prediction. In contrast, \easier achieves more accurate segmentation, camera pose estimation, and 4D reconstruction, resulting in renderings with better visual quality.

\label{sec:addtional_results}

\end{document}